\documentclass[11pt,a4paper]{article}
\usepackage[hyperref]{emnlp2020}
\usepackage{times}
\usepackage{latexsym}

\usepackage{microtype}

\usepackage{bbm}
\usepackage{mathtools}
\usepackage{amsmath}

\usepackage{array}
\usepackage{multirow}
\usepackage{subfiles}
\usepackage{graphicx}
\usepackage{caption}
\usepackage{subcaption}
\usepackage{enumitem}

\usepackage{pdflscape}
\usepackage{amsfonts}

\newcommand\jb[1]{\textcolor{blue}{[JB: #1]}}
\newcommand\mh[1]{\textcolor{purple}{[MH: #1]}}
\newcommand\todo[1]{\textcolor{red}{[TODO: #1]}}

\newcommand\sO{\mathcal{O}}
\newcommand\sV{\mathcal{V}}
\newcommand\sD{\mathcal{D}}
\newcommand{\hh}{\mathbf{h}}

\newcommand\op[1]{\texttt{\MakeUppercase{#1}}}
\newcommand\prop[1]{\textit{#1}}
\newcommand\argn[1]{\texttt{#1}}
\newcommand\dep[1]{\texttt{#1}}

\newcommand\commentout[1]{}

\aclfinalcopy 

\title{Question Decomposition with Dependency Graphs}

\author{Matan Hasson \\
  Tel Aviv University \\
  \texttt{matanhasson@mail.tau.ac.il} \\\And
  Jonathan Berant \\
  Tel Aviv University \\
  The Allen Institue for AI \\
  \texttt{joberant@cs.tau.ac.il} \\}

\date{}

\begin{document}
\maketitle

\begin{abstract}
QDMR is a meaning representation for complex questions, which decomposes questions into a sequence of atomic steps.
While state-of-the-art QDMR parsers use the common sequence-to-sequence (seq2seq) approach, a QDMR structure fundamentally describes labeled relations between spans in the input question, and thus dependency-based approaches seem appropriate for this task.
In this work, we present a QDMR parser that is based on \emph{dependency graphs (DGs)}, where nodes in the graph are words and edges describe logical relations that correspond to the different computation steps. We propose (a) a non-autoregressive graph parser, where all graph edges are computed simultaneously, and (b) a seq2seq parser that uses gold graph as auxiliary supervision.
We find that a graph parser leads to a moderate reduction in performance (0.47$\rightarrow$0.44), but to a 16x speed-up in inference time due to the non-autoregressive nature of the parser, and to improved sample complexity compared to a seq2seq model. Second, a seq2seq model trained with auxiliary graph supervision has better generalization to new domains compared to a seq2seq model, and also performs better on questions with long sequences of computation steps.

\commentout{
\todo{add inference time, success of long decomposition}
In this work we propose a dependencies graph representation for question decomposition, where the dependencies reflect the execution roles of the question tokens within the desired decomposition.
We compare graph-based models with seq2seq-based models on BREAK (a recent dataset annotates a question with the requisite steps for computing its answer), showing the graph-parser has better sample complexity \todo{numbers?}. We introduce \emph{Latent-RAT encoder} architecture for combining these two aspects in a multitask manner, where representations of the relations between the tokens are embedded into the seq2seq encoder but are also used to directly predict the dependencies between the tokens. The novel architecture preserves the seq2seq performances, leverage it with better domain generalization \todo{numbers?}. \mh{graph head...}
\mh{before talking about numbers} For evaluation, we provide a new metric on top of a logical form representation of the decomposition. Besides allowing as to compare our DG to the natural-language decomposition, this metric is more robust to changes that do not affect the semantics of the question.
}
\end{abstract}
\section{Introduction}

\begin{figure}[t]
    \centering
    \includegraphics[width=\linewidth]{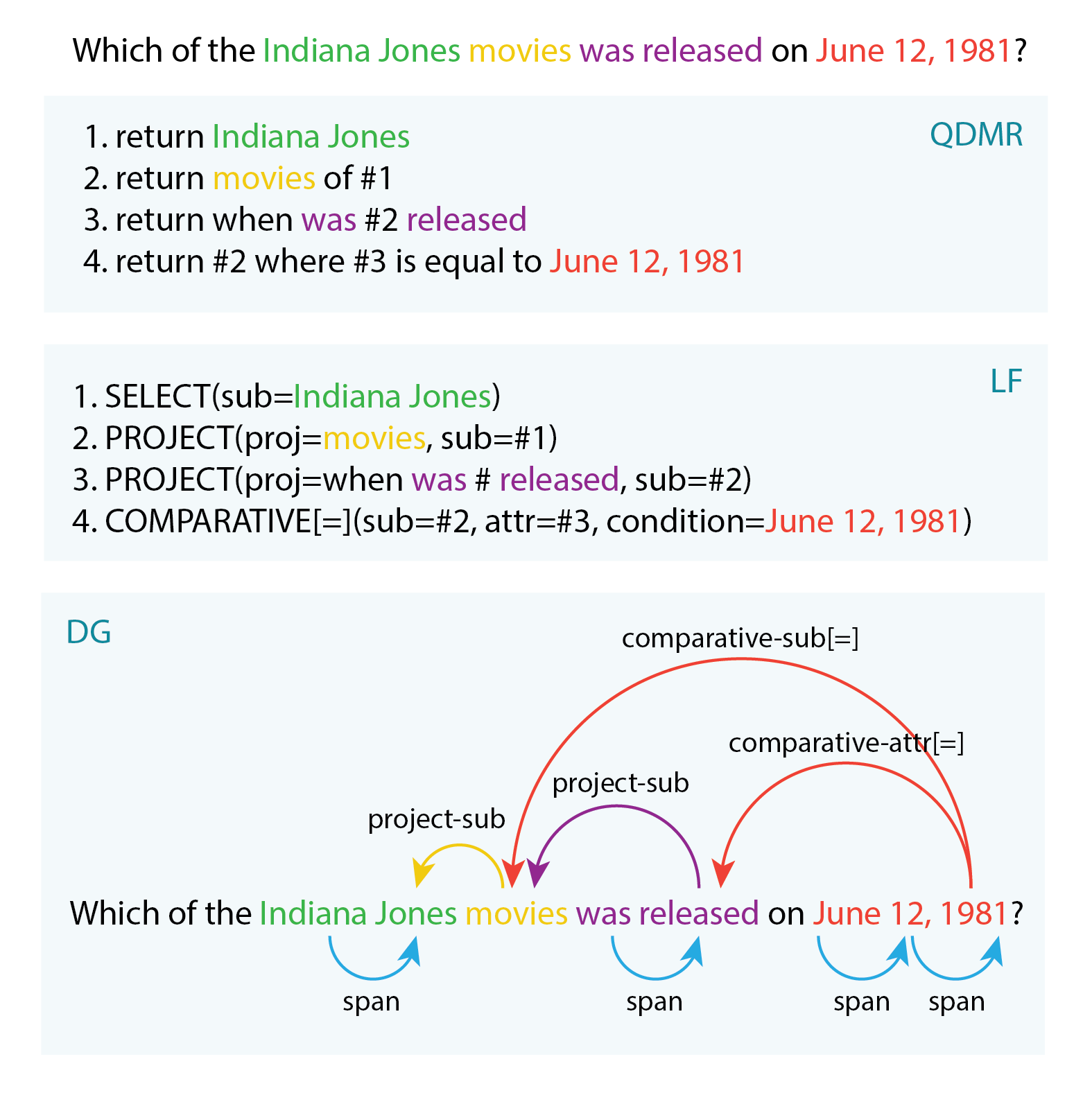}
    \caption{An example question with its corresponding QDMR structure (top), dependency graph over the question tokens (bottom), and an intermediate logical form (LF) used for the QDMR$\rightarrow$DG conversion and for evaluation.}
    \label{fig:example-QDMR-LF-DG}
\end{figure}

Training neural networks to reason over multiple parts of their inputs across modalities such as text, tables, and images, has been a focal point of interest in recent years \cite{Antol_2015_ICCV,pasupat2015compositional,johnson2017clevr,suhr2018corpus,welbl2018constructing,talmor0218web,yang2018hotpotqa,hudson2019gqa,dua2019drop,chen2020hybridqa,hannan2020manymodalqa,talmor2021multimodalqa}. The most common way to check whether a model is capable of complex reasoning, is to pose in natural language a complex question, which requires performing multiple steps of computation over the input.

\commentout{
Question answering (QA), in which a complex natural language question is posed and should be answered given a particular context (text, table, image, etc.), has been rapidly researched for a while. Reasoning and integration of information from multiple parts of the input were applied for various tasks, including images \cite{Antol_2015_ICCV,johnson2017clevr,suhr2018corpus, hudson2019gqa}, paragraphs \cite{dua2019drop}, documents \cite{welbl2018constructing, talmor0218web, yang2018hotpotqa}, tables \cite{pasupat2015compositional} and more.
}

To address the need for better understanding of complex questions, \newcite{wolfson2020break} recently proposed QDMR, a meaning representation where complex questions are represented through a sequence of simpler atomic executable steps (see Fig.~\ref{fig:example-QDMR-LF-DG}), and the final answer is the answer to the final step. QDMR has been shown to be useful for multi-hop question answering (QA) \cite{wolfson2020break} and also for improving interpretability in visual QA \cite{subramanian2020interpretability}.

State-of-the-art QDMR parsers use the typical sequence-to-sequence (seq2seq) approach. However, it is natural to think of QDMR as a dependency graph over the input question tokens. Consider the example in Fig.~\ref{fig:example-QDMR-LF-DG}. The first QDMR step selects the span \emph{``Indiana Jones''}. Then, the next step uses a \texttt{PROJECT} operation to find the \emph{``movies''} of Indiana Jones, and then next step uses another \texttt{PROJECT} operation to find the date when the movies were \emph{``released''}. Such relations can naturally be represented as labeled edges over the relevant question tokens, as shown in Fig.~\ref{fig:example-QDMR-LF-DG}, bottom.

In this work, we propose to use the dependency graph view of QDMR to improve QDMR parsing. We describe a conversion procedure that automatically maps QDMR structures into dependency graphs, using a structured intermediate logical form representation (Fig~\ref{fig:example-QDMR-LF-DG}, middle).
Once we have graph supervision for every example, we train a dependency graph parser, in the spirit of \newcite{dozat2018simpler}, where we predict a labeled relation for every pair of question tokens, representing the logical relation between the tokens. Unlike seq2seq models, this is a non-autoregressive parser, which decodes the entire output structure in a single step.

A second method to exploit the graph supervision is to train a seq2seq model, but have an auxiliary loss term where the graph is decoded from the encoder representations. Towards that end, we propose a Latent-RAT encoder, which uses relation-aware transformer \cite{shaw2018selfattention} to explicitly represent the relation between every pair of input tokens.
Relation-aware transformer has been shown to be useful for encoding graph structures in the context of semantic parsing \cite{wang2020ratsql}.

Last, we propose a new evaluation metric, LF-EM, for QDMR parsing, which is based on the aforementioned intermediate logical form, and show it correlates better with human judgements compared to existing metrics. 

We find that our graph parser leads to a small reduction in LF-EM compared to seq2seq models (0.47$\rightarrow$0.44), but is 16x faster due to its non-autoregressive nature. Moreover, our graph parser has better sample complexity and outperforms the seq2seq model when trained on 10\% of the data or less. When training a seq2seq model along with the auxiliary graph supervision, we find that the parser obtains similar performance when trained on the entire dataset (0.471 LF-EM), but substantially improves performance when generalizing to new domains. Moreover, The Latent-RAT parser performs better on examples with a large number of computation steps.

Our code is available at \url{https://github.com/matanhasson/qdecomp_with_dependency_graphs }.

\section{Overview}
\label{sec:overview}

\begin{figure}
    \centering
    \includegraphics[width=\linewidth]{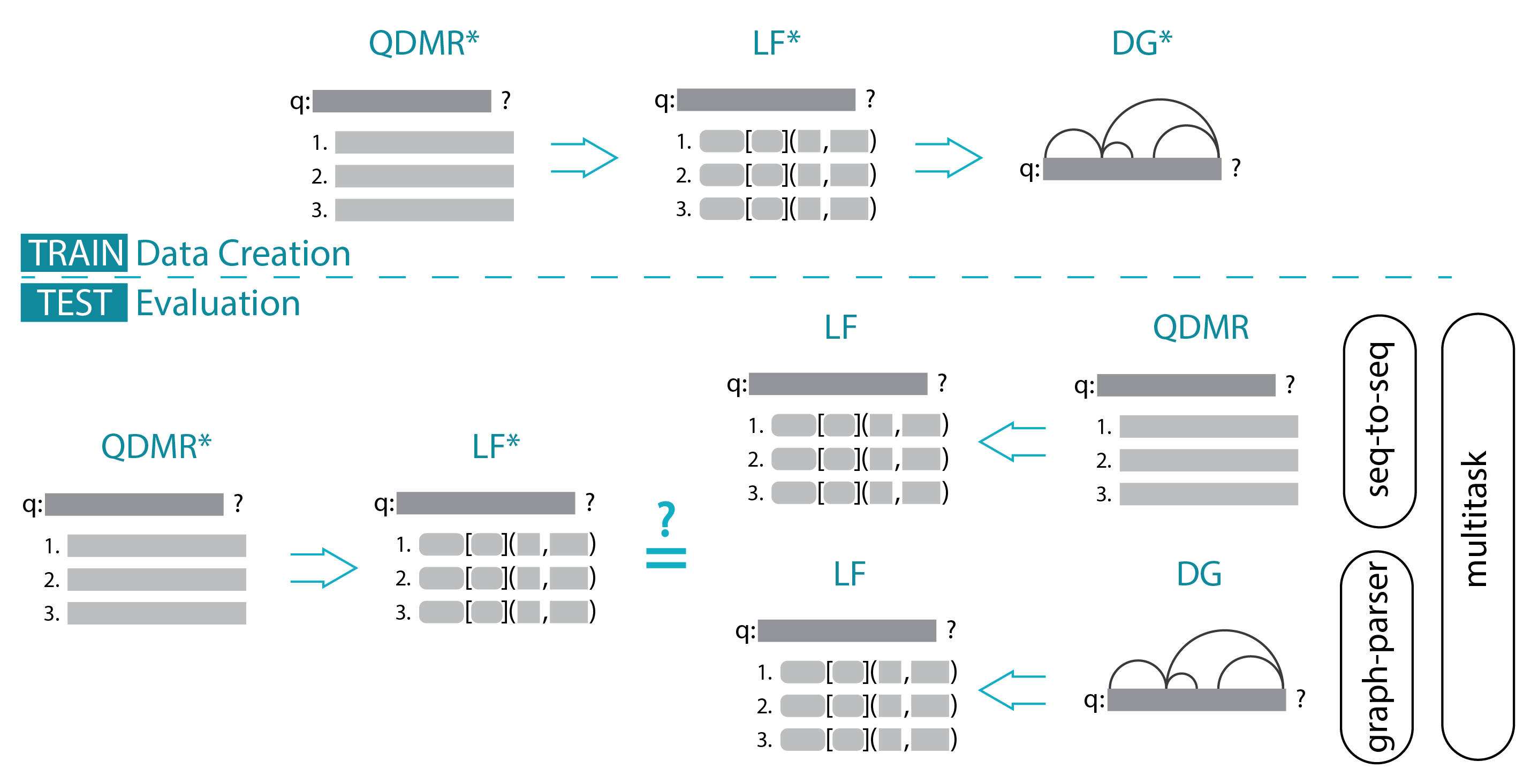}
    \caption{.Overview. For training (top), we create gold DGs from gold QDMRs (\S\ref{sec:graph-creation}) through a conversion into LFs (\S\ref{sec:QDMR-to-LF}). At test time (bottom), we convert model predictions, either QDMRs or DGs, into  LFs (\S\ref{sec:QDMR-to-LF}, \S\ref{sec:DG-to-LF}), and evaluate by comparing them to the gold LFs. Asterisk (*) denotes gold representations.
    }
    \label{fig:overview}
\end{figure}

The core of this work is to examine the utility of a dependency graph (DG) representation for QDMR. We propose conversion procedures that enable training and evaluating with DGs (see Fig.~\ref{fig:overview}). First, we convert gold QDMR structures into logical forms (LF), where each computation step in QDMR is represented with a formal operator, properties and arguments (\S\ref{sec:logical-form:LF-defenition}). Then, we obtain gold DGs by projecting the logical forms onto the question tokens (\S\ref{sec:LF-to-DG}). Once we have question-DG pairs, we can train a graph parser. At test time, QDMRs and DGs are converted into LFs for evaluation. We propose a new evaluation metric over LFs (\S\ref{sec:evaluation}), and show it is more robust to semantic-preserving changes compared to prior metrics.

Our proposed parsers are in \S\ref{sec:models}. On top of standard seq2seq models, we describe (a) a graph parser, and (b) a multi-task model, where the encoder of the seq2seq model is also trained to predict the DG.

\section{QDMR Logical Forms}
\label{sec:logical-form}

QDMR \cite{wolfson2020break} is a text-based meaning representation focused on representing the meaning of complex questions. It is based on a decomposition of questions into a sequence of simpler executable \emph{steps} (Fig. \ref{fig:example-QDMR-LF-DG}), where each step corresponds to a SQL-inspired operator  (Table \ref{tab:logical-form-props-and-args} in Appendix \S\ref{apdx:LF}). 
We briefly review QDMR and then define a logical form (LF) representation based on these operations. We use the LFs both for mapping QDMRs to DGs, and also to fairly evaluate the output of parsers that output either QDMRs directly or DGs.


\subsection{Definitions}
\label{sec:logical-form:LF-defenition}

\paragraph{QDMR Definition}
Given a question with $n$ tokens, $q=\langle q_1, \dots, q_n \rangle$, its QDMR is a sequence of $m$ steps $s^1, \dots, s^m$, where step $s^i$ conceptually maps to a single logical operator $o^i \in \sO$. A step, $s^i$, is a sequence of $n_i$ tokens $s^i=s^i_1, \dots s^i_{n_i}$, where token $s^i_j$ is either a question token $\in \sV_q$ (or some inflection of it), a word from a constant predefined lexicon $\in \sV_{\text{const}}$, or a \emph{reference token} $\in \sV^i_{\text{ref}}=\{\#1, \dots, \#(i-1)\}$, referring to a previous step.
Fig. \ref{fig:QDMR-example-vocabularies} shows an example for a question and its QDMR structure.

\begin{figure}
    \small
    \emph{``Which group from the census is smaller: Pacific islander or African American?''}
    \newline\newline
    1. return census groups                 \\
    2. return \#1 that is Pacific islander   \\
    3. return \#1 that is African American   \\
    4. return size of \#2                    \\
    5. return size of \#3                    \\
    6. return which is lowest of \#4 ,  \#5  
    \begin{equation*}
    \begin{split}
     & \sV_q = \{\dots \textit{group, \textbf{groups},} \dots \textit{small, smaller, smallest,} \dots \} \\
     & \sV^5_{\text{ref}} =\{\textit{\#1, \#2, \textbf{\#3}, \#4}\} \\
     &\sV_{\text{const}} = \sV_{\textit{op}} \cup \sV_{\textit{store}} \cup \sV_{\textit{aux}} \\ 
     &\sV_{\textit{op}} = \{\textit{difference, sum, \textbf{lowest}, highest, for each,}\dots \} \\
     &\sV_{\textit{store}} = \{\textit{population, \textbf{size}, elevation, flights, price, date}\dots \} \\
     &\sV_{\textit{aux}} = \{\textit{a, is, are, \textbf{of}, \textbf{that}, the, with, was, did, to}\dots \}
    \end{split}
    \end{equation*}

    \caption{
    QDMR annotation vocabularies. Each example is annotated with a lexicon that consists of: $\sV_q$, the question tokens and their inflections; $\sV^i_{\text{ref}}$, references to previous steps; $\sV_{\text{const}}$, constant terms including operational terms ($\sV_{\textit{op}}$), domain-specific  words that are not in the question, such as \emph{size} ($\sV_{\textit{store}}$); and auxiliary words like prepositions ($\sV_{\textit{aux}}$). Boldface indicates words used in the QDMR structure.
    }
    \label{fig:QDMR-example-vocabularies}
\end{figure}

\paragraph{QDMR Logical Form (LF)}
Given a QDMR $S = \langle q ; s^1, \dots , s^m \rangle$, its \emph{logical form} is a sequence of logical form steps $Z =  \langle q; z^1, \dots ,z^m \rangle$. The LF step $z^i$, corresponding to $s^i$ , is a triplet $z_i = \langle o^i, \rho^i, A^i \rangle$ where $o^i \in \sO$ is the logical operator; $\rho^i \in \textit{PROP}(o^i)$ are operator-specific properties; and $A^i$ is a dictionary of arguments, mapping an operator-specific argument $\eta \in \textit{ARG}(o^i)$ to a span $\tau$ from the QDMR step $s^i$.
For convenience, we denote $z^i$ with the string $o^i[\rho^i](\eta^i_1=\tau^i_1, \dots)$.
Table~\ref{tab:logical-form-props-and-args_short} provides a few examples for the mapping from QDMR to LF steps, and 
Table \ref{tab:logical-form-props-and-args} in the Appendix provides the full list.

Imposing more structure on QDMR through the LF is useful (a) for evaluation (\S\ref{sec:evaluation}), where LFs detect differenet QDMR phrasings that have identical meaning, and (b) for creating DGs, since the operators, properties and arguments, will be used to define the labels of edges in the DG (see Fig~\ref{fig:example-QDMR-LF-DG}).

\begin{table}
\small
\centering
\begin{tabular}{ m{1.5cm} m{0.5cm} m{1cm} m{3cm} }
\hline \textbf{Operator} & \textbf{PROP} & \textbf{ARG} & \textbf{Example} \\ \hline
\op{select}      & $\emptyset$         & \argn{sub}               & 
return cubes \\ &&& 
\op{SELECT}[](\argn{sub}=cubes) \\  
\hline
\op{filter}      & $\emptyset$         & \multirow{2}{1cm}{\argn{sub, cond}}         & 
return \#1 from Toronto \\ &&&
\op{FILTER}[](\argn{sub}=\#1, \argn{cond}=from Toronto) \\ 
\hline
\op{aggregate}   & \multirow{5}{0.5cm}{\prop{max, min, count, sum, avg}}      & \argn{arg}               &
return maximal number of \#1    \\&&&
\op{AGGREGATE}[\prop{max}](\\&&& \argn{arg}=\#1)   \\ \\ \\
\hline
\op{arithmetic}   & \multirow{4}{0.5cm}{\prop{sum, diff, mult, div}}      &  \multirow{3}{1cm}{\argn{arg, left, right}}               &
return the difference of \#3 and \#4    \\&&&
\op{ARITHMETIC}[\prop{diff}]( \\&&& \argn{left}=\#3, \argn{right}=\#4)   \\ \\
\hline
\end{tabular}
\caption{\label{tab:logical-form-props-and-args_short} LF operators, properties and arguments (partial list,  see Table~\ref{tab:logical-form-props-and-args} for full list). 
}
\end{table}

\paragraph{QDMR$\rightarrow$LF}
\label{sec:QDMR-to-LF}
We convert QDMRs to LFs with a rule-based method, extending the procedure for detecting operators from \citet{wolfson2020break} to also find properties and arguments. To detect properties we use a lexicon, see details in Table~\ref{tab:logical-form-props-indicators} (Appendix \S\ref{apdx:LF}), and in our public implementation.


\subsection{LF-based Evaluation (LF-EM)}
\label{sec:evaluation}

The official evaluation metric for QDMR\footnote{\url{https://leaderboard.allenai.org/break/submissions/public}} is normalized EM (NormEM), where the predicted and gold QDMR structures are normalized using a rule-based procedure, and then exact string match is computed between the two normalized QDMRs. Since in this work we convert both QDMRs (\S\ref{sec:logical-form:LF-defenition}) and DGs (\S\ref{sec:DG-to-LF}) to LFs, we propose LF-EM, a LF-based evaluation metric, and show that it correlates better with notions of semantic equivalence.

\commentout{
\citet{wolfson2020break} supplied multiple metrics for evaluating QDMRs, but none of them taking into account the structure induced by the operators . The official evaluation metric, Normalized Exact Match \mh{appendix? say its prior work?}, tries to normalize the decomposition in a rule-based manner by farther breaking it down to even smaller components, normalize each component and reorder them in a consistent way. However, this metric assumes the steps to be in a proper fluent English due to the usage of POS tagging.
Our predicted dependencies graphs may add or subtract information that mostly is logically unnecessary, but "break" the whole phrase in terms of POS parsers. Here we provide an evaluation metric that is based on the QDMR LF (\S\ref{sec:logical-form:LF-defenition}). Therefore, it is agnostic to the specific representation of the QDMR, and focus on logical common representation allowing us comparing different representation of QDMR. Leveraged with our specification of operators, properties and named arguments, it tend to capture logical equivalence better, and is more consistent with manual evaluation \jb{should we empirically show this somewhere?}. Note we show the transformations from native QDMR and dependencies graph to LF in \S\ref{sec:QDMR-to-LF} and \S\ref{sec:DG-to-LF} respectively. \\
}

LF-EM essentially involves computing exact match between the predicted and gold LFs, using the LF described above. To further capture semantic equivalences, we perform a few more normalization steps. We briefly describe these normalizations, and give the full description in \S\ref{sec:evaluation-LF-EM}. 

\begin{figure}
    \centering
    \includegraphics[width=\linewidth]{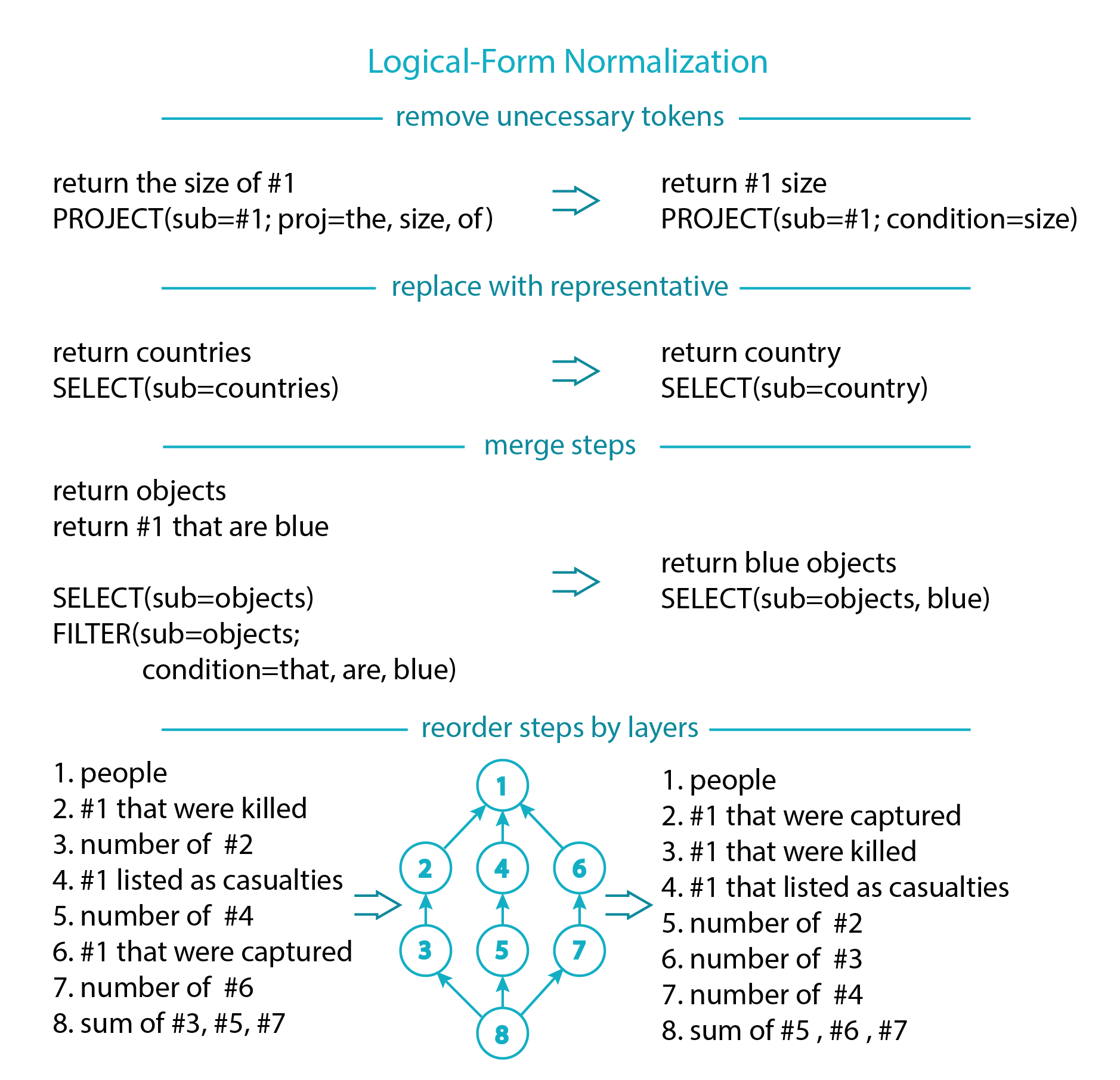}
    \caption{An illustration of LF normalization. Normalization is done on the LF $Z$, and we present QDMR steps for ease of reading.
    } 
    \label{fig:eval-normalization}
\end{figure}

Given a logical form $Z$, we transform each step to a normalized form, and the final textual representation is given by representing each step as described in \S\ref{sec:logical-form}: \op{operator}[\prop{property}](\argn{arg}=\dots; \dots).
We apply the following steps (Fig.~\ref{fig:eval-normalization}): 
\commentout{(1) 
\emph{Remove} unnecessary tokens; (2) \emph{Normalize} tokens; (2) \emph{Merge} steps ; and (3) \emph{Reorder} steps. 
In the end, a textual representation is given by representing each step with the convention described in \S\ref{sec:logical-form}: \textit{operator[property](arg=\dots; \dots)}. 
}

\paragraph{Remove and normalize tokens}
Each LF step includes a list of tokens in its arguments. In this normalization step, we remove lexical items, such as \emph{``max''}, which  are used to detect the operator and property (Table~\ref{tab:logical-form-props-indicators} in \S\ref{apdx:LF}), as those are already represented outside the arguments. In addition, we remove words from a stop word list ($\sV_{aux}$, see Fig.~\ref{fig:QDMR-example-vocabularies}). Finally, we use a synonym list to represent words in such a list with a single representative (\emph{countries}$\rightarrow$\emph{country}). 

\commentout{
Our LF steps are already pretty normalized in terms of operator, properties and the names of the arguments; so all we have left is to normalize in terms of the arguments tokens. We remove \emph{operator indicators} and \emph{property indicators} (Table \ref{tab:logical-form-props-indicators}) since this information is already embedded in the operator and the properties them-self. Recall each QDMR sample was annotated with a specific lexicon (\S\ref{sec:logical-form}, Figure \ref{fig:QDMR-example-vocabularies}). We remove auxilirary tokens, $\sV_{aux}$, that were added in the first place to allow continuous fluent sentences to be written. Finally, we replace each token with its \emph{representative token} by defining equivalence classes of \emph{break equivalent tokens} (Table \ref{tab:break-equivalance}) and set a representative for each class.
}

\paragraph{Merge Steps} 
QDMR annotations sometime vary in their granularity. For example one example might contain \emph{`return metal objects'}, while another might have \emph{`return objects; return \#1 that are metal'}. This is especially common in \op{FILTER} and \op{PROJECT} steps. We merge chains of \op{filter} steps, as well as \op{filter} or \op{project} steps that follow a \op{select} step. See details in \S\ref{sec:eval-logical-form:merge-rules}.


\paragraph{Reorder steps}
QDMR describes a directed acyclic graph of computation steps, and there are multiply ways to order the steps (Fig. \ref{fig:eval-normalization}). We recursively compute the \emph{layer} of each step as $\text{layer}(s) = \max{\{\text{layer}(s^{\text{ref}}_1), \dots \}} + 1$, where the maximization is over all steps $s$ refers to. We then re-order steps by layer and then lexicographically.

We manually evaluate the metrics normalized EM and LF-EM on 50 random development set examples using predictions from the CopyNet-BERT model (see \S\ref{sec:experiments}). We find that both (binary) metrics have perfect precision: they only assign credit when indeed the QDMR reflects the correct question decomposition, as judged by the authors. However, LF-EM covers more examples, where the LF-EM on this sample is 52.0, while normalized EM is 40.0. Thus, LF-EM provides a tighter lower bound on the performance of a QDMR parser and is a better metric for QDMR parsing. 

\section{From LFs to Dependency Graphs}
\label{sec:graph-creation}
\label{sec:LF-to-DG}
\label{sec:DG-to-LF}

\begin{figure*}
    \centering
    \includegraphics[width=\textwidth]{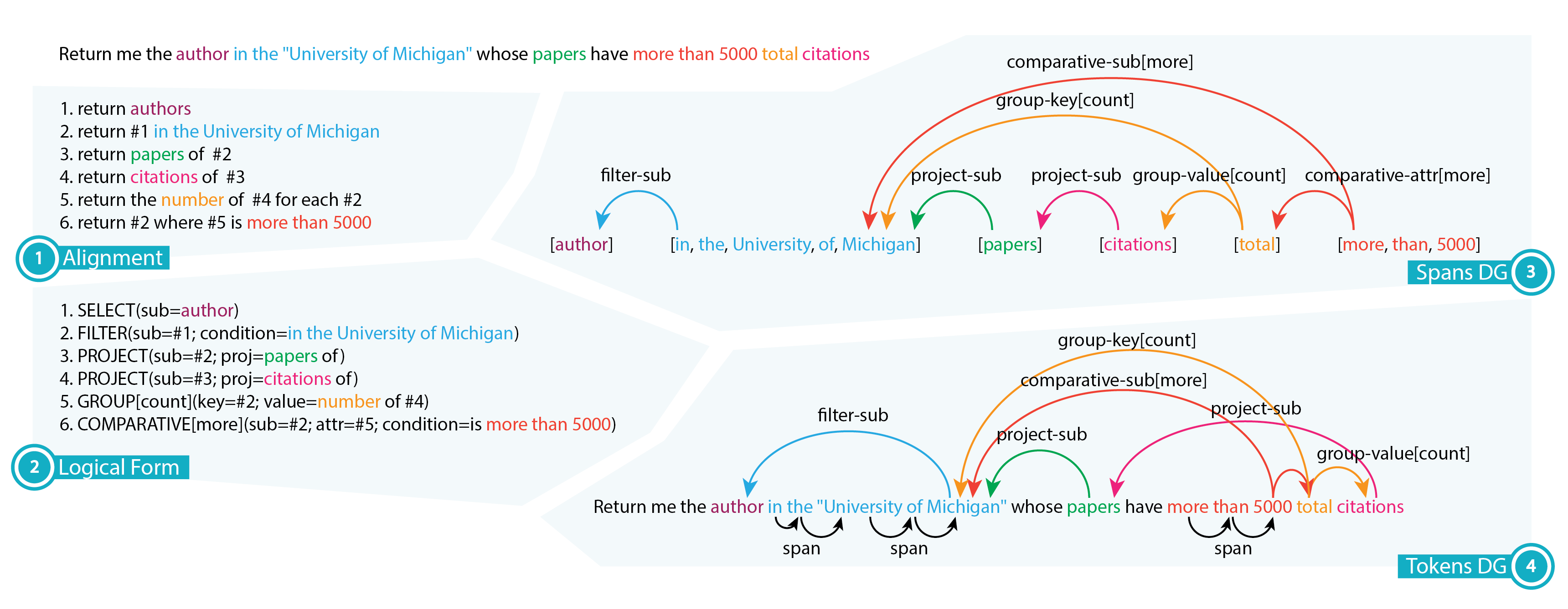}
    \caption{Dependency graph creation: 
    (1) \emph{Token alignment}: align question tokens and QDMR step tokens. 
    (2) \emph{Logical Form}: extract the LF of the QDMR. (3) \emph{SDG extraction}: induced from the LF and the alignment. 
    (4) \emph{DG Creation}: convert the SDG to a DG.}
    \label{fig:graph-creation}
\end{figure*}

Given a QDMR decomposition $S=\langle q;s^1, \dots ,s^m \rangle$, we construct a dependency graph $G= \langle \mathcal{N},\mathcal{E} \rangle$, where the nodes $\mathcal{N}$ correspond to question tokens, and the edges $\mathcal{E}$ describe the logical operations, resulting in a graph with the same meaning as $S$.

\commentout{
\mh{alternative 1:}In high level (Fig. \ref{fig:graph-creation}) , we first align the steps tokens to the input question tokens. Then, we connect the tokens that are aligned to the same step with \emph{span} dependencies; and induce from the QDMR-LF the dependencies between the spans. However, since we would like to convert the DG back to LF for evaluation, this transformation is not always straight forward (\S\ref{sec:graph-creation:challenges}). Thus we add an intermediate data structure, \emph{Spans Dependencies Graph}, which can be thought as a graph representation of the QDMR-LF; where each node is associated with a decomposition step and contains its aligned tokens, and the edges are the dependencies between the steps. The SDG serves as a stable anchor structure, giving us a free hand in dealing with representation issues in a variety of ways. We describe the conversion from SDG to TDG (stands for \emph{Tokens Dependencies Graph}, emphasizes its the graph on top of the question tokens) taking into account the reversibility demand.
}

The LF$\rightarrow$DG procedure is shown in Fig.~\ref{fig:graph-creation} and consists of the following steps:
\begin{itemize}[topsep=-0pt,itemsep=0pt,partopsep=0pt, parsep=0pt]
    \item Token alignment: align each token in the question to a token in a QDMR step.
    \item Spans Dependency Graph (SDG) extraction: construct a graph where each node corresponds to a list of tokens in a QDMR step, and edges describe the dependencies between the steps. 
    \item Dependency Graph (DG) extraction: convert the SDG to a DG over the question tokens. Here, we add \dep{span} edges for tokens that are in the same step, and deal with some representation issues (see~\S\ref{sec:sdg_to_dg}).
\end{itemize}

Because we convert predicted DGs to LFs for evaluation, the LF$\rightarrow$DG conversion must be invertible. We now describe the details of the LF$\rightarrow$DG conversion. Our conversion succeeds in 97.12\% of the BREAK dataset \cite{wolfson2020break}.


\subsection{Token Alignment}
\label{sec:graph-creation:tokens-alignment}

We denote the question tokens by $q=\langle q_1 \dots q_n \rangle$ and the $i$th QDMR step tokens by $\forall i\in[1..m], s^i=s^i_1 \dots s^i_{n_i}$.  An \emph{alignment} is defined by $M=\{(q_i,s^k_j) \mid  q_i\approx s^k_j ;i\in[1..n], k\in[1..m], j\in[1..n_k] \}$, where by $t\approx t'$ we mean $t, t'$ are either identical or equivalent. Roughly speaking, these equivalences are based on BREAK annotation lexicon (Fig. \ref{fig:QDMR-example-vocabularies}) - in particular. the inflections of the question tokens $\sV_q$ (e.g , \emph{``object''} and \emph{``objects''}), and equivalence classes on top of the constant lexicon $\sV_\textit{const}$ (e.g , \emph{``biggest''} and \emph{``longest''}). See Table \ref{tab:break-equivalance} in Appendix \S\ref{sec:evaluation-LF-EM} for more details.

\begin{figure}[ht]
    \centering
    \includegraphics[width=\linewidth]{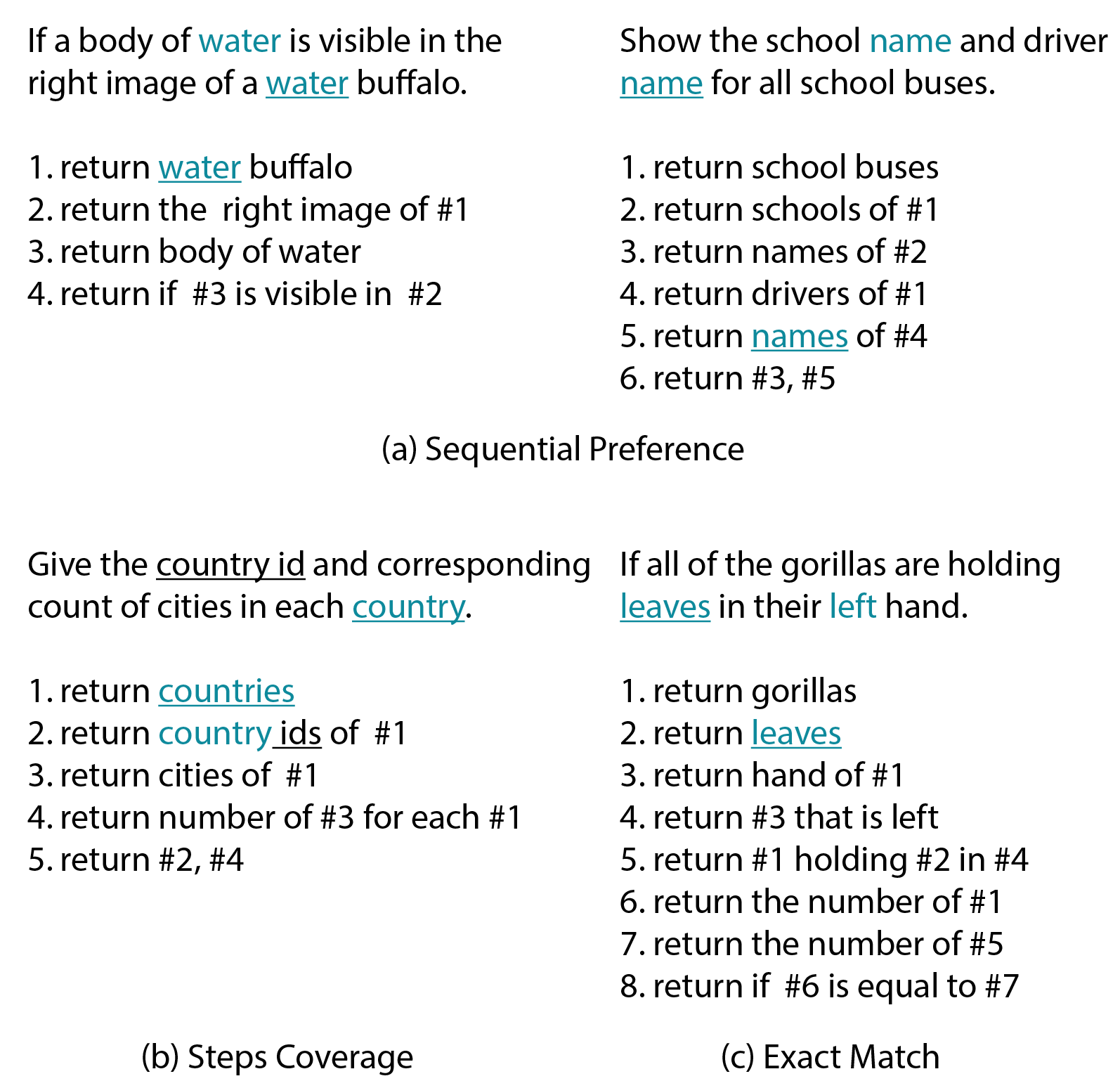}
    \caption{
    Heuristics for token alignment. Potential tokens for alignment colored, where the preferable choice according to the heuristic is underlined. On the top left, the second occurrence of \emph{``water''} is preferred in QDMR step \#1 due to the adjacent word \emph{``buffalo''}. On the top right, the second occurrence of \emph{``name''} is preferred in QDMR step \#5, because this step refers to \#4 that contains the word \emph{``drivers''}.
    }
    \label{fig:graph-creation:tokens-alignment}
\end{figure}

To find the best alignment $M$, we formulate an optimization problem in the form of an Integer Linear Program (ILP) and use a standard ILP solver.\footnote{\url{https://developers.google.com/optimization}} 
The full details are given as a part of our open source implementation.
The objective function uses several heuristics to assign a high score to an alignment the has the following properties (Fig. \ref{fig:graph-creation:tokens-alignment}): 
\begin{itemize}[topsep=-0pt,itemsep=0pt,partopsep=0pt, parsep=0pt]
    \item \emph{Minimalism}: Aligning each question token to at most one QDMR step token and vice versa is preferable.
    \item \emph{Exact Match}: Aligning a question token to a QDMR token that is identical is preferable.
    \item \emph{Sequential Preference}: 
    Aligning long sequences from the question to a single step is preferable (when a step has a reference token (\emph{\#1}), we take into account the tokens in the referenced step, see Fig.~\ref{fig:graph-creation:tokens-alignment}, top right).
    \item \emph{Steps Coverage}: Covering more steps is preferable.
\end{itemize}


\subsection{Spans Dependencies Extraction}
\label{sec:graph-creation:spans-dependencies-extraction}

Given the QDMR, LF, and alignment $M$, we construct the Span Dependency Graph (SDG). Each QDMR step is a node labeled by a list of tokens (spans). The list of tokens is computed with the alignment $M$, where given a QDMR step $s^k$, the list contains all question tokens $q_i$, such that $(q_i, s^k_j) \in M$, where $s^k_j$ is a word in $s^k$. The list is ordered according to the position in the question.

Edges in the SDG are computed using reference tokens. If step $s_i$ has a reference token to step $s_j$, we add an edge $(s_i, s_j)$ (we abuse notation and refer to SDG nodes and QDMR steps with the same notation). Each edge has a \emph{tag}, which is a triple consisting of the operator $o^i$  of the source node $s_i$, the property $\rho^i$ of the source node, and the named argument $\eta^i_{\text{ref}}$ that contains the reference token. For readability we denote the tag triplet $\textit{tag}(i,j) = \langle o^i, \rho^i, \eta^i_\text{ref} \rangle$ by $o^i$-$\eta^i_\text{ref}[\rho^i]$. Figure~\ref{fig:graph-creation} shows an extracted SDG.

\commentout{
Given a QDMR, $\langle q; s^1, \dots, s^m \rangle$, we say that step $s^i$ \emph{depends} on step $s^j$ if $s^i$ refers to $s^j$. The tag of this dependency is induced from the LF representation of the QDMR (\S\ref{sec:logical-form}), by taking into account the operator of step  $s^i$, $o^i$; its operational properties, $\rho^i$, and the name of the argument that refers to $s^j$, $\eta^i_k$ (Fig. \ref{fig:example-QDMR-LF-DG}). In particular, $tag(i,j) = \langle o^i, \rho^i, \eta^i_k \rangle$ and for readability we sometimes write $o^i-\eta^i_k[\rho^i]$. We denote by $\sD$ the possible \emph{operational-role dependencies} tags, i.e $\sD=\{\langle o, \rho, \eta \rangle \mid o \in \sO, \rho \in \textit{PROP}_o, \eta \in \textit{ARG}_o\}$. \emph{Spans Dependencies Graph (SDG)} holds the dependencies between QDMR steps and maps each step to its aligned tokens in the question. Formally, it is defined by $G_{spans}=\langle N, E;  \mathcal{S} \rangle$ where $N=[1 .. m]$ and $E=\{(i,j,t) \mid i,j \in [1..m], t \in \sD, s^i \text{ depends on } s^j, tag(i,j)=t\}=\{(i,j,t) \mid i,j \in [1..m], t \in \sD, \exists k : "\#j" \in \tau^{i,k} \text{ and } t=\langle o^i, \rho^i, \eta^i_k \rangle \}$.  $\mathcal{S}:N\rightarrow [1..n]$ is the mapping function from a step index to an ordered subsequence of the question tokens indices, representing a list of \emph{spans} from the question. Given a \emph{tokens alignment} $M$, we construct the mapping  $\mathcal{S}_M(i) = \langle j\in[1..n] \mid \exists k \ (q_j, s^i_k) \in M \rangle $.
}

\subsection{SDG$\rightarrow$DG}
\label{sec:sdg_to_dg}

We construct a DG by projecting the SDG on the question tokens. This is done by: (a) For each SDG node and its list of tokens, add edges between the tokens from left-to-right with a new \texttt{span} tag (black edges in Fig~\ref{fig:graph-creation}); (b) use the rightmost word in every span as its representative for the edges between different spans.

\begin{figure}
    \centering
    \includegraphics[width=\linewidth]{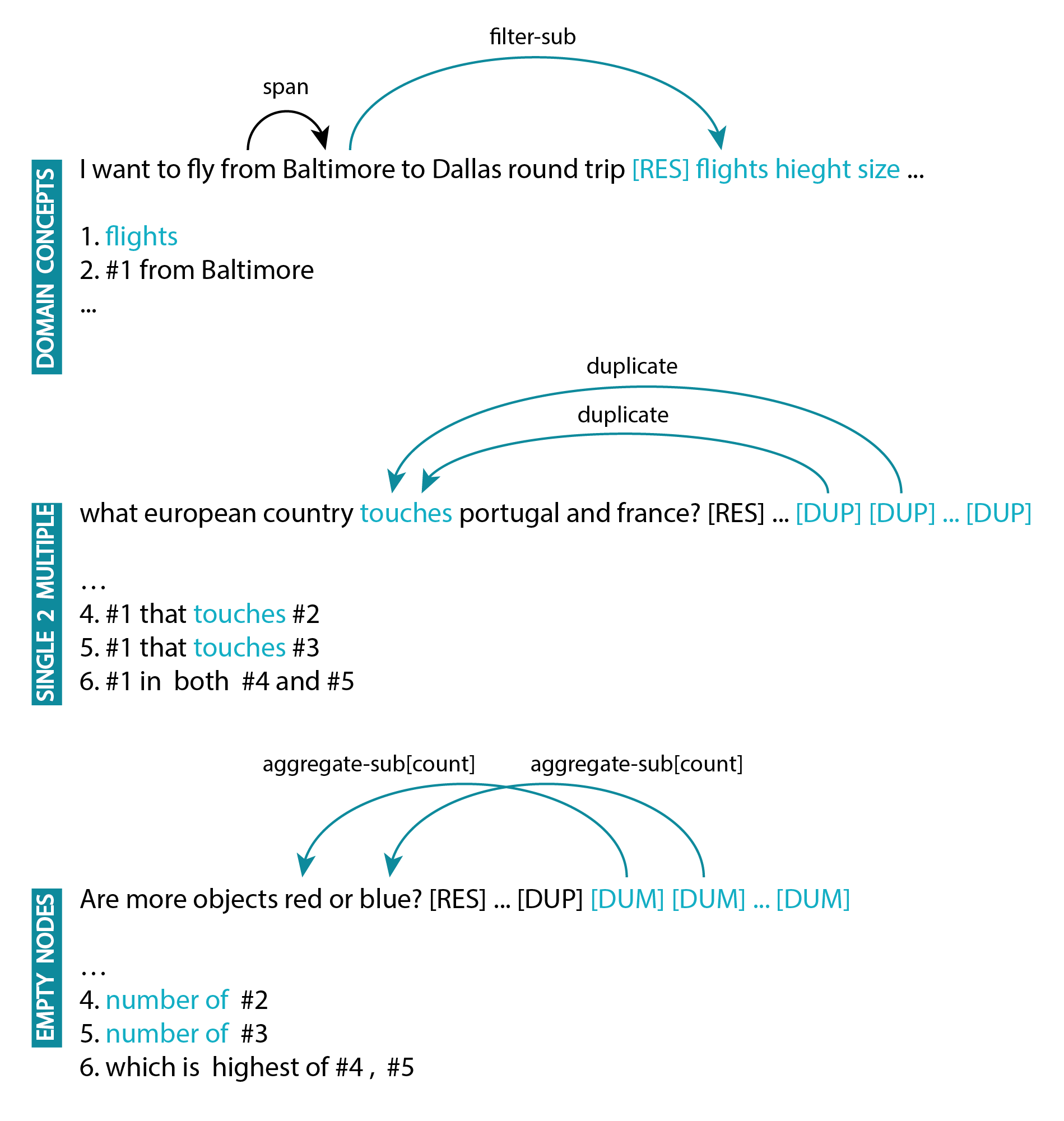}
    \caption{Representation Issues. Projecting the SDG over the question token (DG) is not always trivial. We solve this by concatenating special tokens to the question.}
    \label{fig:dependencies-graph:representation-issues}
\end{figure}

\commentout{
Given a question $q=q_1 \dots q_n$, its \emph{Dependencies Graph (TDG)} is a directed graph $G\langle N, E \rangle$ where the nodes are the question tokens and the edges are labeled with operational-role dependencies from $\sD \cup \{\dep{span}\}$ (\S\ref{sec:graph-creation:spans-dependencies-extraction}), i.e $N=[1..n]$, $E=\{(i,j,t) \mid i,j \in N, t \in \sD \cup \{\dep{span}\} \}$. The \dep{span} tag groups the tokens into steps. Each token is allowed to be a part of at most a single span, so we can convert TDG back to SDG for evaluation.

We construct TDG from SDG by simply (1) connect tokens belonging to the same span with \dep{span}, left to right; (2) refer the rightmost token of a span as its representative and induces the SDG dependencies between theses representatives.
}
However, this transformation is non-trivial for two reasons. First, some SDG nodes do not align to any question token, Second, some question tokens align to multiple SDG nodes, which does not allow the DG to be converted back to an SDG unambiguously for evaluation.
We now explain how we resolve such representation issues,  mostly based on adding more tokens to the input question.
\commentout{
\subsection{Representation Challenges}
\label{sec:graph-creation:challenges}
TDGs supply a structured representation of a QDMR's LF over the question tokens. In order to reconstruct the LF representation (i.e its equivalent SDG), TDG must cover all the relevant information and be unambiguous in terms of deterministic conversion into LF. In this section we elaborate some of the main issues, and the way we overcome them for proper reconstruction. 
}
Fig. \ref{fig:dependencies-graph:representation-issues} illustrates the different types of challenges and our proposed solution.

\paragraph{Domain-specific concepts}
QDMR annotators were allowed to use a small number of tokens that are pragmatically assumed to exist in the domain ($\sV_\textit{store}$ in Fig.~\ref{fig:QDMR-example-vocabularies}). For example, when annotating ATIS questions \cite{atis-dataset}, the word \emph{``flight''} is allowed to be used in the QDMR structure even if it does not appear in the question, since this is a flight-reservation domain. We concatenate all the words in $\sV_\textit{store}$ to the end of each question after a special separator token, which allows token alignment (\S\ref{sec:graph-creation:tokens-alignment}) to map such QDMR steps to a question word (Fig.~\ref{fig:dependencies-graph:representation-issues}, top).

\commentout{
QDMR steps where annotated using a dedicated vocabulary, which extends the question tokens set with some auxiliary tokens $\sV_\textit{const}$ (Fig. \ref{fig:QDMR-example-vocabularies}). We distinguish between the \emph{operational} tokens $\sV_\textit{op}$ (e.g 'not', 'for each', 'maximum'), that are already embedded in the operator and the properties, and \emph{resources} tokens $\sV_\textit{store}$ (e.g 'flights', 'size', 'date'), that are mostly domain oriented and designed to compensate for the lack of information in the question. To allow TDG to refer $\sV_\textit{store}$ tokens, we concatenate them to the question, right after the special token $\textit{[RES]}$, so the \emph{tokens alignment} process (\S\ref{sec:graph-creation:tokens-alignment}) takes them into account.
}

\paragraph{Empty SDG nodes}
some steps only contains tokens that are not in the question (e.g, \emph{``Number of \#2''} in Fig.~\ref{fig:dependencies-graph:representation-issues} bottom), and thus their list of tokens in the SDG node is empty. In this case, we cannot ground the SDG node in the question. Therefore we add a constant number of \emph{dummy tokens}, \emph{[DUM]}, which are used to ground such SDG nodes.

\paragraph{Single tokens to multiple SDG nodes}
A single question token can be aligned to multiple SDG nodes.
Recall the tokens of each SDG nodes are connected with a chain of \texttt{span} edges. This leads to cases where two chains that pass through the same question token cannot be distinguished when converting the DG back to an SDG for evaluation.
We solve this by concatenating a constant number of special $\textit{[DUP]}$ tokens that conceptually duplicate another token by referring to it with a new \dep{duplicate} tag. 
Now, each span chain uses a different copy of the shared token by referring to the $\textit{[DUP]}$ instead of the original one.


\commentout{
\mh{Thesis only?}Note we tried another approach for solving the discussed issues based on enriched dependencies with no additional dummy tokens, which their role might be hard for learning. We skipped empty nodes (with no aligned tokens) by referring their references, and concatenate the dependencies across the path to track it. Since this technique tend to be harder for decoding (deterministic unwind concatenated path, ILP) we focused on the additional tokens based approach.
}

\section{Models}
\label{sec:models}

Once we have methods to convert QDMRs to DGs and LFs, and DGs to LFs, we can evaluate the advantages and disadvantages of standard autoregressive decoders compared to graph-based parsers. We describe three models: (a) An autoregressive parser,  (b) a graph parser, (c) an autoregressive parser that is trained jointly with a graph parser in a multi-task setup. For a fair comparison, all models have the same BERT-based encoder \cite{devlin2019bert}.

\paragraph{CopyNet-BERT (baseline)}
This autoregressive QDMR parser is based on  the CopyNet baseline from \citet{wolfson2020break}, except we replace the BiLSTM encoder with a transformer initialized with BERT. The model encodes the question $q$ and then decodes the QDMR $S$ step-by-step and token-by-token.

The QDMR decoder is an LSTM \cite{hochreiter1997long} augmented with a copy mechanism \cite{gu2016incorporating}, where at each time step the model either decodes a token from the vocabulary or a token from the input. Since the input is tokenized with word pieces, we average word pieces that belong to a single word to get word representations, which enables word copying. Training is dones with standard maximum likelihood.


\paragraph{Biaffine Graph Parser (BiaffineGP)}
The biaffine graph parser takes as input the quesiton $q$ augmented with the special tokens described in \S\ref{sec:sdg_to_dg} and predicts the DG by classifying for every pair of tokens whether there is an edge between them and the label of the edge. The model is based on the biaffine dependency parser of \newcite{dozat2018simpler}, except here we predict a \emph{DAG} and not a tree, so each node can have more that one outgoing edge.

Let $\mathbf{H} = \langle \hh_1, \dots, \hh_{|\mathbf{H}|} \rangle$ be the sequence of representations output by the BERT encoder. The biaffine parser uses four 1-hidden layer feed-forward networks over each contextualized token representation  $\hh_l$: 
\begin{align*}
\hh_l^{\text{edge-head}} = \textit{FF}^{\text{edge-head}}(\hh_l), \\
\hh_l^{\text{edge-dep}} = \textit{FF}^{\text{edge-dep}}(\hh_l), \\ 
\hh_l^{\text{label-head}} = \textit{FF}^{\text{label-head}}(\hh_l), \\
\hh_l^{\text{label-dep}} = \textit{FF}^{\text{label-dep}}(\hh_l).
\end{align*}

The probability of an edge from token $i$ to token $j$ is given by $\sigma(\hh_i^{\text{edge-dep}} W_{\text{edge}} \hh_j^{\text{edge-head}})$, where $W_{\text{edge}}$ is a parameter matrix. Similarly, the score of an edge labeled by the tag $t$ from token $i$ to token $j$ is given by $s_{ij}^t = \hh_i^{\text{label-dep}} W_t \hh_j^{\text{label-head}}$, where $W_t$ is the parameter matrix for this tag.  We then compute a distribution over the set of tags $\mathcal{T}$ with $\text{softmax}(s_{ij}^{1}, \dots, s_{ij}^{|\mathcal{T}|})$. 

Training is done with maximum likelihood both on the edge probabilities and label probabilities. Inference is done by taking all edges with edge probability $>0.5$ and then labeling those edges according to the most probable tag.

\commentout{
\mh{from related work:} It factorized into two modules: one that predicts whether or not a directed edge exists between two tokens, and another that predicts the best label for each potential edge. They concatenate word and POS tag embeddings, and feed them into a multilayer bidirectional LSTM. For each of the two modules, they use single-layer feedforward networks (FNN) to split the top recurrent states into two parts-a head representation and a dependent representation. They then use bilinear or biaffine classifiers which are generalizations of linear classifiers to include multiplicative interactions between two vectors—to predict edges and labels. 
The classifiers produce $n \times n$ and $n \times n \times |T|$ logits matrices correspondingly, where $n$ is the sequence length and $T$ is the tags set. The loss is the summation of the two cross entropy losses. On inference, the overall prediction is given by taking the tag with the maximal probability of arcs with $p_\textit{arc}$ greater from some threshold.

We relay on their architecture, but replace the BiLSTM encoder with a BERT encoder, and average word-pieces encodings for each token.
}

There is no guarantee that the biaffine parser will output a valid DG. For example, if an SDG node has an outgoing edge labeled with \dep{filter-sub} and another labeled with \dep{project-sub}, we cannot tell if the operator is \op{filter} or \op{project}.
This makes parsing fail, which occurs in 1.83\% of the cases. 
To create a SDG, we first use the \dep{span} edges to contruct SDG nodes with lists of tokens, and then add edges between SDG nodes by projecting the edges between tokens to edges between the SDG nodes. To prevent cases where parsing fails, we can optionally apply an ILP that takes the predicted probabilities as input, and outputs a valid DG. 
The exact details are given in our open source implementation.

\commentout{
Note the output graph it not necessarily a valid TDG. We softly convert it to a SDG by first group the tokens into spans according to the \emph{span} dependencies (two tokens are considered in the same span if there is an undirected path of \emph{span} dependencies between them), and than add dependencies between the groups by project the dependencies between their members. Moreover, we provide an ILP based decoding (post processing) in Appendix \ref{sec:gependencies-graph-decoding}. 
\mh{Thesis only: Variations: tags-only, multilabel, Operational Aware}
}

\paragraph{Multi-task Latent-RAT Encoder}

\begin{figure}
    \centering
    \includegraphics[width=\linewidth]{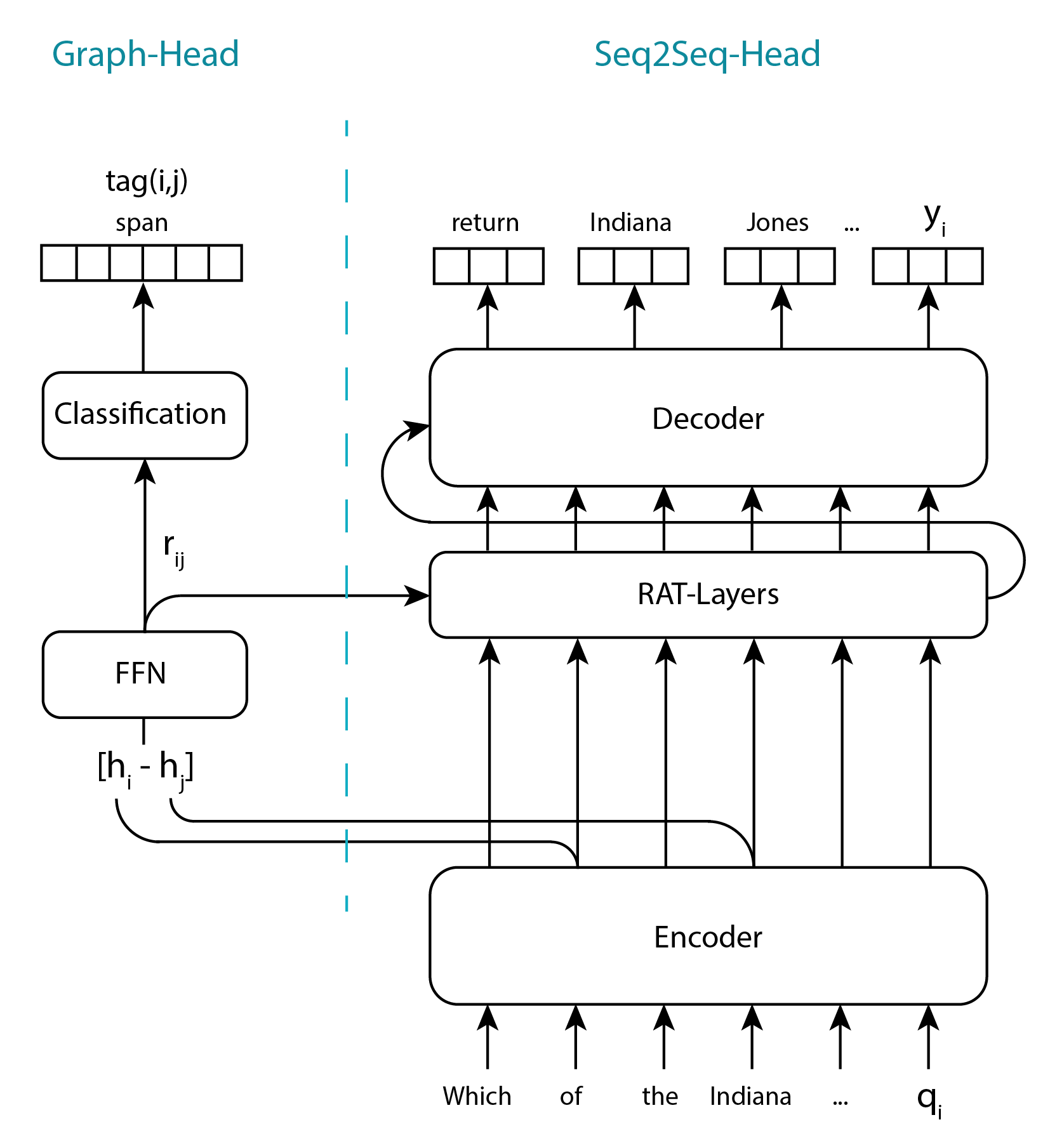}
    \caption{
    Latent-RAT architecture. 
    The encoder hidden states are used to represent the relations between the question tokens, $r_{ij}$. Then, these representations are used for both (1) direct prediction of the dependency (tag) between the tokens (graph-head); and (2) augment the encodings via RAT layers (seq2seq-head).
    The sub-networks for $r^K_{ij}$, $r^V_{i,j}$
    are symetric, and represented by the $r_{ij}$.}
    \label{fig:models:latent-rat-architecture}
\end{figure}
In this model, our goal is to improve the sequence-to-sequence parser by providing more information to the encoder using the DG supervision. Our model will take the question $q$ (with special tokens as before) as input, and predict both the graph $G$ directly and the QDMR structure $S$ with a decoder.

We would like the information on relations between tokens to be part of the transformer encoder, unlike the biaffine parser that uses separate feed-forward networks for that, so that the decoder can take advantage of this information. To accomplish that, we use RAT transformer layers
\cite{shaw2018selfattention, wang2020ratsql}, which explicitly represent relations between tokens, and have been shown to be useful for encoding graphs over input tokens in the context of semantic parsing.

RAT layers inject information on the relation between tokens inside the transformer self-attention mechanism \cite{vaswani2017attention}. Specifically, the similarity score $e_{ij}$ computed using queries and keys is given by:
$$e_{ij} \propto \hh_iW_Q(\hh_jW_K+\textcolor{red}{r^K_{ij}})^T,$$where $W_Q, W_K$ are the query and key parameter matrices and the only change is the term $r^K_{ij}$, which represents the relation between the tokens $i$ and $j$. Similarly, the relation between tokens is also considered when summing over self-attention values: $$\sum_{j=1}^{\mathbf{H}}{\alpha_{ij}(x_jW_V+\textcolor{red}{r^V_{ij}})},$$ where $W_V$ is the value parameter matrix, $\alpha_{ij}$ is the attention distribution and the only change is the term $r^V_{ij}$.

\commentout{
\mh{from related work:}
The main idea of the RAT architecture is based on adding relative position information in a self-attention layer \cite{shaw2018self} as follows:
\begin{equation}
\small
\begin{split}
    e^{(h)}_{ij}&=\frac{x_iW^{(h)}_Q(x_jW^{(h)}_K+\textcolor{red}{r^K_{ij}})^T}{\sqrt{d_z/H}} \\
    z^{(h)}_i&=\sum_{j=1}^n{\alpha^{(h)}_{ij}(x_jW^{(h)}_V+\textcolor{red}{r^V_{ij}})}
\end{split}
\end{equation}

Where the $r_{ij}$ terms encode the relationship between the input elements $x_i, x_j$. The rest is the standard self-attention encoder, i.e Transformer \cite{vaswani2017attention}.
}
Unlike prior work where the terms $r_{ij}^K, r_{ij}^V$ were learned parameters, here we want these vectors to (a) be a function of the contextualized representation and (b) be informative for classifying the dependency label in the gold graph. By learning latent representations from which the gold graph can be decoded, we will provide useful information for the sequence-to-sequence decoder. Specifically, given a RAT layer with representations $\hh_i, \hh_j$ for tokens $i$ and $j$, we represent relations and compute a loss in the following way (see Figure~\ref{fig:models:latent-rat-architecture}):
\begin{align*}
    r_{ij}^K &= FF^K(\hh_i - \hh_j), \\
    S^K &= R^KW^{\text{out}} + b^K \in \mathbbm{R}^{n \times n \times |\mathcal{T}|},\\
    \textit{Loss}^K &= \textit{CE}(S^K).
\end{align*}
$FF^K$ is a 1-hidden layer feed-forward network, $R^K \in \mathbb{R}^{(n \times n) \times d_{\text{transformer}}}$ is a concatenation of all $r_{ij}^K$ for all pairs of tokens, $W^{\text{out}} \in \mathbb{R}^{d_\text{transformer} \times |\mathcal{T}|}$ is a projection matrix that provides a score for all possible labels (including the NONE label).

We compute an analogous loss $\textit{Loss}^V$for $r_{ij}^V$ and the final graph loss is $\textit{Loss}^K + \textit{Loss}^V$ over all RAT layers. To summarize, by performing multi-task training with this graph loss we push the transformer to learn representations $r_{ij}$ that are informative of the gold graph, and can then be used by the decoder to output better QDMR structures.

\commentout{
In contrast to the SQL setup in which the schema is given in advanced, the gold relations (QDMR dependencies) are unknown in our setup. Therefore we build the relation representations as a function of the pairwise base (seq2seq) encoder representations, and demand it could by classified to the gold dependency by a  classification layer. Conceptually, instead of encoding a given relation and inject it into the transformer, we assume such relation exists in a latent manner, encode it based on the base encoder representation and ensure it indeed encoded the right dependency. Thus we get an auxiliary loss for the seq2seq head and a by product graph parser head. Note we extend the dependencies tag with 'NONE' tag, stands for no-dependency.

\begin{equation*}
\small
\begin{split}
& e = \textit{Encoder(x)} \\
& r^K_{i,j} = FF^K(e_i-e_j) \\
& r^V_{i,j} = FF^V(e_i-e_j) \\
& S^K = R^KW^K + b^K \in \mathbbm{R}^{n \times n \times T}\\
& S^V = R^VW^V + b^V \in \mathbbm{R}^{n \times n \times T} \\
&\textit{Loss}_{graph} = CE(S^K)+CE(S^V) \\
&\textit{Loss} = \gamma \textit{Loss}_{s2s} + (1-\gamma)\textit{Loss}_{graph}
\end{split}
\end{equation*}

Where $r^K_{ij}$, $r^V_{ij}$ are the relations representation for the RAT layer. For graph inference, we multiply the probabilities induced by the logits $S^K, S^V$:
\begin{equation*}
\small
\begin{split}
& p^K(i,j,t)=\textit{softmax}\limits_t S^K_{i,j,t} \\
& p^V(i,j,t)=\textit{softmax}\limits_t S^V_{i,j,t} \\
& p(i,j,t)=p^K(i,j,t)p^V(i,j,t) \\
& y = \{(i,j,t) \mid  t=\textit{argmax}_{t} p(i,j,t)\}
\end{split}
\end{equation*}
}
\section{Experiments}
\label{sec:experiments}

We now describe our empirical evaluation of the models described above.

\subsection{Experimental Setup}
We build our models in AllenNLP \cite{Gardner2017AllenNLP}, and use BERT-base \cite{devlin2019bert} to produce contextualized token representations in all models. We train with the Adam optimizer \cite{kingma2017adam}. Our Latent-RAT model include 4 RAT layers, each with 8 heads.
Full details on hyperparameters and training procedure in Appendix \S\ref{apdx:experiments-parameters}.

We examine the performance of our models in three setups:
\begin{itemize}[leftmargin=*,itemsep=0pt,topsep=0pt]
    \item \emph{Standard}: We use the official BREAK dataset.
    \item \emph{Sample Complexity (SC)}: We examine the performance of models with decreasing amounts of training data. The goal is to test which model has better sample complexity.
    \item \emph{Domain Generalization (DomGen)}: We train on 7 out of 8 sub-domains in BREAK and test on the remaining domain, for each target domain. The goal is to test which model generalizes better to new domains.
\end{itemize}

As an evaluation metric, we use LF-EM and also the official BREAK metric, normalized EM, when reporting test results on BREAK.

\subsection{Results}


\paragraph{Standard setup}
Table \ref{tab:results:full-train} compares the performance of the different models (\S\ref{sec:models}) to each other and to the top entries on the BREAK leaderboard. As expected, initializing CopyNet with BERT dramatically improves test performance (0.388$\rightarrow$0.47). The Latent-RAT sequence-to-sequence model achieves similar performance (0.471), and the biaffine graph parser, BiaffineGP, is slightly behind with an LF-EM of 0.44 (but has faster inference, as we show below). Adding an ILP layer on top of BiaffineGP to eliminate constraint violations in the output graph improves performance to 0.454.

While our proposed models do not significantly improve performance in the LF-EM setup, we will see next that they improve domain generalization and sample complexity. Moreover, since BiaffineGP is a non-autoregressive model that predicts all output edges simultaneously, it dramatically reduces inference time.

Last, the top entry on the BREAK leaderboard uses BART \cite{lewis2019bart}, a pre-trained seq2seq model (we use a pre-trained encoder only), which leads to a state-of-the-art LF-EM of 0.496.

\begin{table}
\small
\begin{tabular}{lllrl}
\hline
Model                          & \multicolumn{2}{c}{NormEM}                              & \multicolumn{2}{c}{LF-EM}                          \\
                               & \multicolumn{1}{c}{dev}    & \multicolumn{1}{c}{test}   & \multicolumn{1}{c}{dev} & \multicolumn{1}{c}{test} \\ \hline \hline
CopyNet                        & -                          & \multicolumn{1}{r}{0.294}  & \multicolumn{1}{l}{-}   & 0.388                        \\
BART$_\textit{leaderboard \#1}$ & -                          & \multicolumn{1}{r}{0.389} & \multicolumn{1}{l}{-}   & 0.496                  \\ \hline
CopyNet+BERT                   & \multicolumn{1}{r}{0.373}  &  0.375                          & 0.474                   & 0.47                         \\
BiaffineGP                & -                          & -                          & 0.441                   & 0.44                         \\
BiaffineGP$_\textit{ILP}$ & -                          & -                          & \multicolumn{1}{l}{0.453}    & 0.454                         \\
Latent-RAT & \multicolumn{1}{r}{0.356} &  0.363                          & 0.469                   &  0.471                        \\
\hline
\end{tabular}
\caption{
\label{tab:results:full-train}
Normalized EM and LF-EM on the development and test sets of BREAK.
}
\end{table}

\paragraph{Domain generalization}
Table \ref{tab:results:domain-generalization} shows LF-EM on each of BREAK's sub-domains when training on the entire dataset (top), when training on all domains but the target domain (middle), and the relative drop compared to the standard setup (bottom). 

The performance of BiaffineGP and Latent-RAT is higher compared to CopyNET+BERT in the DomGen setup. In particular, the performance of Latent-RAT is the best in 7 out of 8 sub-domains, and the performance of BiaffineGP is the best in the last domain. Moreover, Latent-RAT outperforms CopyNet+BERT in all sub-domains. We also observe that the performance drop is lower for BiaffineGP and Latent-RAT compared to CopyNet+BERT. Overall, this shows that using graphs as a source of supervision leads to better domain generalization.


\begin{table*}
\small
\centering
\begin{tabular}{lrrrrrrrr}
\hline
Model            & \multicolumn{1}{l}{ATIS} & \multicolumn{1}{l}{CLEVR} & \multicolumn{1}{l}{COMQA} & \multicolumn{1}{l}{CWQ} & \multicolumn{1}{l}{DROP} & \multicolumn{1}{l}{GEO} & \multicolumn{1}{l}{NLVR2} & \multicolumn{1}{l}{SPIDER} \\ \hline \hline
CopyNet+BERT & 0.58              & \textbf{0.564}    & 0.562             & \textbf{0.36}     & 0.473             & \textbf{0.66}    & 0.344             & 0.369             \\
BiaffineGP   & \textbf{0.591}    & 0.489             & 0.595             & 0.322             & 0.445             & 0.62             & 0.293             & \textbf{0.41}     \\
Latent-RAT    & 0.589             & 0.524             & \textbf{0.598}    & 0.316             & \textbf{0.479}    & 0.64             & \textbf{0.353}    & 0.376             \\ \hline
CopyNet+BERT & 0.282             & 0.351             & 0.423             & 0.173             & 0.131             & 0.52             & 0.039             & 0.189             \\
BiaffineGP   & 0.302             & 0.339             & \textbf{0.483}    & 0.168             & 0.146             & 0.52             & 0.04              & 0.197             \\
Latent-RAT    & \textbf{0.335}    & \textbf{0.356}    & 0.435             & \textbf{0.189}    & \textbf{0.149}    & \textbf{0.58}    & \textbf{0.063}    & \textbf{0.201}    \\ \hline
CopyNet+BERT & -51.38\%          & -37.77\%          & -24.73\%          & -51.94\%          & -72.30\%          & -21.21\%         & -88.66\%          & -48.78\%          \\
BiaffineGP   & -48.90\%          & \textbf{-30.67\%} & \textbf{-18.82\%} & -47.83\%          & \textbf{-67.19\%} & -16.13\%         & -86.35\%          & -51.95\%          \\
Latent-RAT    & \textbf{-43.12\%} & -32.06\%          & -27.26\%          & \textbf{-40.19\%} & -68.89\%          & \textbf{-9.38\%} & \textbf{-82.15\%} & \textbf{-46.54\%} \\ \hline
\end{tabular}
\caption{
\label{tab:results:domain-generalization}
Domain Generalization. LF-EM on the development set per sub-domain when training on the entire training set (top), and when training on all domains except the target one (middle). The bottom section is the performance drop from the full setup to the DomGen setup. 
}
\end{table*}

\paragraph{Sample Complexity}
Table~\ref{tab:results:sample-complexity} shows model performance as a function of the size of the training data. While the LF-EM of BiaffineGP is lower given the full training set (Table~\ref{tab:results:full-train}), when the size of the training data is small it substantially outperforms other models, improving performance by 3-4 LF-EM points given 1\%-10\% of the data. With 20\%-50\% of the data Latent-RAT and CopyNet+BERT have comparable performance.


\begin{table}[]
\small
\centering
\begin{tabular}{lrrrrr}
\hline
Model            & 1\%            & 5\%            & 10\%           & 20\%           & 50\%           \\ \hline \hline
CopyNet$_\textit{BERT}$     & 0.112          & 0.261          & 0.323          & 0.38           & 0.426          \\
BiaffineGP & \textbf{0.159} & \textbf{0.296} & \textbf{0.351} & 0.382          & 0.411          \\
Latent-RAT    & 0.003          & 0.227          & 0.326          & \textbf{0.383} & \textbf{0.432} \\ \hline
\end{tabular}
\caption{
\label{tab:results:sample-complexity}
Development set LF-EM as a function of the size of the training set.
}
\end{table}

\paragraph{Inference time for the graph parser}
The graph parser, BiaffineGP, is a non-autoregressive model that predicts all output edges simultaneously, as opposed to a sequence-to-sequence model that decodes a single token at each step. We measure the average runtime of the forward pass for both BiaffineGP and CopyNet+BERT and find that BiaffineGP has an average runtime of 0.08 seconds, compared to 1.306 seconds of CopyNet+BERT -- a 16x speed-up.

\subsection{Analysis}
\label{sec:analysis}

\begin{figure}
    \centering
    \includegraphics[width=\linewidth]{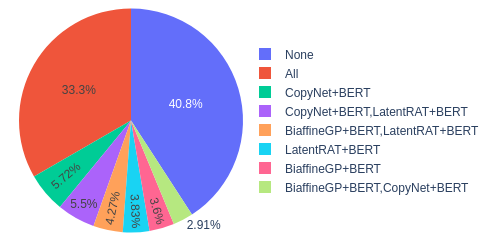}
    \caption{Model agreement in terms of LF-EM on the development set. Each slice gives the fraction of examples predicted correctly by a subset of models. }
    \label{fig:analysis:overlap}
\end{figure}

\paragraph{Model agreement}
Figure~\ref{fig:analysis:overlap} shows model agreement between CopyNet+BERT, BiaffineGP, and Latent-RAT on the development set. 
Roughly 60\% of the examples are predicted correctly by one of the models, indicating that enembling the three models could result in further performance improvement.

The agreement of Latent-RAT with CopyNet+BERT (5.5\%) and BiaffineGP (4.27\%) is greather than the overlap between CopyNet+BERT and BiaffineGP, perhaps since it is a hybrid of a seq2seq and graph parser. Moreover, in 3.83\% of the examples, Latent-RAT is the only model with a correct prediction.

\begin{figure}
    \centering
    \includegraphics[width=\linewidth]{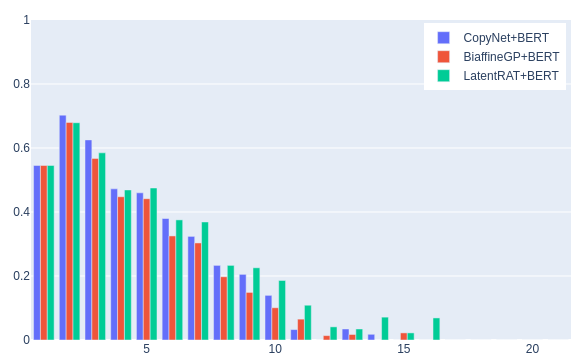}
    \caption{LF-EM on the development set per number of steps. We compute for each example its (gold) number of steps, and calculate the average LF-EM per bin.}
    \label{fig:analysis:LF-EM-per-num-of-steps}
\end{figure}

\paragraph{Length analysis}
We compared the average LF-EM of models for each possible number of steps in the QDMR structure
(Fig. \ref{fig:analysis:LF-EM-per-num-of-steps}). 
We observe that CopyNet+BERT outperforms Latent-RAT when the number of steps is small, but once the number of steps is $\geq 5$, Latent-RAT outperforms CopyNet+BERT, showing it is handles complex decompositions better, and in agreement with the tendency of sequence-to-sequence models to struggle with long output sequences.

\paragraph{Error analysis}
\label{sec:analysis:error-analysis}
We randomly sampled 30 errors from each model and manually analyzed them. Table \ref{tab:error-analysis-percent} describes the error classes for each model, and Appendix \ref{apdx:error-analysis} provides examples for these classes. Each example can have more than one error cateogry.

For all models, the largest error category is actually cases where the prediction is correct but not captured by the LF-EM metric: 70\% for CopyNet+BERT, 40\% for BiaffineGP and 73.3\% for Latent-RAT. This shows that the performance of current QDMR parsers is actually quite good, but capturing this with an automatic evaluation is challenging. Cases where the output is correct include:
\begin{itemize}[leftmargin=*,itemsep=0pt,topsep=0pt]
    \item \emph{Equivalent Solutions}: the prediction is logically equivalent to the gold structure. 
    \item \emph{Elaboration Level}: the model prediction is more/less granular compared to the gold structure, but the prediction is correct.
    \item \emph{Redundancy}: additional information is predicted/omitted that does not effect the computation. For example the second occurrence of \emph{``yard's'} in , \textit{"2. return \textbf{yards} of \#1; 3. \#1 where \#2 is lower than 10 \textbf{yards}"}. 
    \item \emph{Wrong Gold} - cases where the predication is more accurate than the gold decomposition.
\end{itemize}
The main classes of errors are:
\begin{itemize}[leftmargin=*,itemsep=0pt,topsep=0pt]
    \item \emph{Missing Information}: missing steps, missing references or missing tokens that affect the result of the computation.
    \item \emph{Additional Steps}:  duplicate steps or additional steps that change the result of the computation. 
    \item \emph{Wrong Global Structure}: The computation described by the predicted structure is wrong (for example, addition instead of subtraction).
    \item \emph{Wrong Step Structure}: incoherent structure of a particular step that cannot be mapped to a proper structure. 
    \item \emph{Out of Vocabulary}: seq2seq models sometimes predict tokens that are not related to the question nor the decomposition. For example, \textit{"rodents"} in a question about flowers. 
\end{itemize}

\begin{table}[]
\small
\begin{tabular}{llll}
\hline                      & CopyN$_\textit{BERT}$ & BiaGP & LatRAT \\ \hline \hline
Correct               & 70.00\%      & 40.00\%    & 73.30\%    \\ \hline
Equivalent Solutions & 40.00\%      & 30.00\%    & 43.30\%    \\
Elaboration Level     & 26.70\%      & 23.30\%    & 26.70\%    \\
Redundancy            & 10.00\%      & 0.00\%     & 3.30\%     \\
Wrong Gold            & 0.00\%       & 3.30\%     & 3.30\%     \\ \hline
Missing Information   & 16.70\%      & 20.00\%    & 10.00\%    \\
Additional Steps      & 6.70\%       & 23.30\%    & 6.70\%     \\
Wrong Global Structure           & 3.30\%       & 16.70\%    & 13.30\%    \\
Wrong Step Structure  & 0.00\%       & 16.70\%    & 6.70\%     \\
Out of Vocab          & 13.30\%      & 0.00\%     & 0.00\%     \\ \hline
\end{tabular}
\caption{
\label{tab:error-analysis-percent}
Error classes and their frequency over a sample of 30 random errors.
Model names were shortened from CopyNet+BERT, BiaffineGP and Latent-RAT.
}
\end{table}
\section{Conclusion}

In this work, we propose to represent QDMR structures with a dependency graph over the input tokens, and propose a graph parser and a seq2seq model that uses graph supervision as an auxiliary loss. We show that a graph parser is 16x faster than a seq2seq model, and that it exhibits better sample coplexity. Moreover, using graphs as auxiliary supervision improves out-of-domain generalization and leads to better performance on questions that represent a long sequence of computational steps. Last, we propose a new evaluation metric for QDMR parsing and show it better corresponds to human intuitions.

In future work, we will examine ensemble models that take into account the complementary nature of graph parsers and seq2seq parser to further improve performance on QDMR parsing.

\commentout{
\todo{add inference time, success of long decomposition}
In this work we provide a dependency graph representation for questions decomposition, based on the QDMR formalism. We examine sequence-to-sequence based model against graph parser and show the later one has better sample complexity, as well as lower performances drop rate on domain generalization. We utilize these generalization features and present the Latent-RAT encoder - an encoder that integrates graph based supervision into the a sequence encoder. Our Latent-RAT model  gets the best performances when training on all the subdatasets of BREAK but one (termed leave-one-out setup), and also improves the sequence-to-sequence sample complexity. 
\\
In addition, we supply a logical form representation - a more structural SQL-like representation of the QDMR- and a new evaluation metric on top of it. It allows us comparing different representation of the same decomposition, as well as more robust to variants of semantic equivalent decompositions.

\subsection{Future work}
Evaluation might include logical equivalence for alternative solutions, either by templates equivalence classes or by a synthetic execution framework.
}

\section*{Acknowledgments}
We thank Vivek Kumar Singh for his helpful ILP guidelines, and Tomer Wolfson for having kindly assisted in running our evaluation metric on our predictions for BREAK test set. This research was partially supported by The Yandex Initiative for Machine Learning, and the European Research Council (ERC) under the European
Union Horizons 2020 research and innovation programme (grant ERC DELPHI 802800).


\bibliography{emnlp2020new}

\begin{thebibliography}{25}
\expandafter\ifx\csname natexlab\endcsname\relax\def\natexlab#1{#1}\fi

\bibitem[{Antol et~al.(2015)Antol, Agrawal, Lu, Mitchell, Batra, Zitnick, and
  Parikh}]{Antol_2015_ICCV}
Stanislaw Antol, Aishwarya Agrawal, Jiasen Lu, Margaret Mitchell, Dhruv Batra,
  C.~Lawrence Zitnick, and Devi Parikh. 2015.
\newblock \href {https://doi.org/10.1109/ICCV.2015.279} {{VQA:} visual question
  answering}.
\newblock In \emph{2015 {IEEE} International Conference on Computer Vision,
  {ICCV} 2015, Santiago, Chile, December 7-13, 2015}, pages 2425--2433. {IEEE}
  Computer Society.

\bibitem[{Chen et~al.(2020)Chen, Zha, Chen, Xiong, Wang, and
  Wang}]{chen2020hybridqa}
Wenhu Chen, Hanwen Zha, Zhiyu Chen, Wenhan Xiong, Hong Wang, and William~Yang
  Wang. 2020.
\newblock \href {https://doi.org/10.18653/v1/2020.findings-emnlp.91}
  {{H}ybrid{QA}: A dataset of multi-hop question answering over tabular and
  textual data}.
\newblock In \emph{Findings of the Association for Computational Linguistics:
  EMNLP 2020}, pages 1026--1036, Online. Association for Computational
  Linguistics.

\bibitem[{Devlin et~al.(2019)Devlin, Chang, Lee, and
  Toutanova}]{devlin2019bert}
Jacob Devlin, Ming-Wei Chang, Kenton Lee, and Kristina Toutanova. 2019.
\newblock \href {https://doi.org/10.18653/v1/N19-1423} {{BERT}: Pre-training of
  deep bidirectional transformers for language understanding}.
\newblock In \emph{Proceedings of the 2019 Conference of the North {A}merican
  Chapter of the Association for Computational Linguistics: Human Language
  Technologies, Volume 1 (Long and Short Papers)}, pages 4171--4186,
  Minneapolis, Minnesota. Association for Computational Linguistics.

\bibitem[{Dozat and Manning(2018)}]{dozat2018simpler}
Timothy Dozat and Christopher~D. Manning. 2018.
\newblock \href {https://doi.org/10.18653/v1/P18-2077} {Simpler but more
  accurate semantic dependency parsing}.
\newblock In \emph{Proceedings of the 56th Annual Meeting of the Association
  for Computational Linguistics (Volume 2: Short Papers)}, pages 484--490,
  Melbourne, Australia. Association for Computational Linguistics.

\bibitem[{Dua et~al.(2019)Dua, Wang, Dasigi, Stanovsky, Singh, and
  Gardner}]{dua2019drop}
Dheeru Dua, Yizhong Wang, Pradeep Dasigi, Gabriel Stanovsky, Sameer Singh, and
  Matt Gardner. 2019.
\newblock \href {https://doi.org/10.18653/v1/N19-1246} {{DROP}: A reading
  comprehension benchmark requiring discrete reasoning over paragraphs}.
\newblock In \emph{Proceedings of the 2019 Conference of the North {A}merican
  Chapter of the Association for Computational Linguistics: Human Language
  Technologies, Volume 1 (Long and Short Papers)}, pages 2368--2378,
  Minneapolis, Minnesota. Association for Computational Linguistics.

\bibitem[{Gardner et~al.(2018)Gardner, Grus, Neumann, Tafjord, Dasigi, Liu,
  Peters, Schmitz, and Zettlemoyer}]{Gardner2017AllenNLP}
Matt Gardner, Joel Grus, Mark Neumann, Oyvind Tafjord, Pradeep Dasigi,
  Nelson~F. Liu, Matthew Peters, Michael Schmitz, and Luke Zettlemoyer. 2018.
\newblock \href {https://doi.org/10.18653/v1/W18-2501} {{A}llen{NLP}: A deep
  semantic natural language processing platform}.
\newblock In \emph{Proceedings of Workshop for {NLP} Open Source Software
  ({NLP}-{OSS})}, pages 1--6, Melbourne, Australia. Association for
  Computational Linguistics.

\bibitem[{Gu et~al.(2016)Gu, Lu, Li, and Li}]{gu2016incorporating}
Jiatao Gu, Zhengdong Lu, Hang Li, and Victor~O.K. Li. 2016.
\newblock \href {https://doi.org/10.18653/v1/P16-1154} {Incorporating copying
  mechanism in sequence-to-sequence learning}.
\newblock In \emph{Proceedings of the 54th Annual Meeting of the Association
  for Computational Linguistics (Volume 1: Long Papers)}, pages 1631--1640,
  Berlin, Germany. Association for Computational Linguistics.

\bibitem[{Hannan et~al.(2020)Hannan, Jain, and Bansal}]{hannan2020manymodalqa}
Darryl Hannan, Akshay Jain, and Mohit Bansal. 2020.
\newblock \href {https://aaai.org/ojs/index.php/AAAI/article/view/6294}
  {Manymodalqa: Modality disambiguation and {QA} over diverse inputs}.
\newblock In \emph{The Thirty-Fourth {AAAI} Conference on Artificial
  Intelligence, {AAAI} 2020, The Thirty-Second Innovative Applications of
  Artificial Intelligence Conference, {IAAI} 2020, The Tenth {AAAI} Symposium
  on Educational Advances in Artificial Intelligence, {EAAI} 2020, New York,
  NY, USA, February 7-12, 2020}, pages 7879--7886. {AAAI} Press.

\bibitem[{Hemphill et~al.(1990)Hemphill, Godfrey, and
  Doddington}]{atis-dataset}
Charles~T. Hemphill, John~J. Godfrey, and George~R. Doddington. 1990.
\newblock \href {https://www.aclweb.org/anthology/H90-1021} {The {ATIS} spoken
  language systems pilot corpus}.
\newblock In \emph{Speech and Natural Language: Proceedings of a Workshop Held
  at Hidden Valley, {P}ennsylvania, June 24-27,1990}.

\bibitem[{Hochreiter and Schmidhuber(1997)}]{hochreiter1997long}
Sepp Hochreiter and J{\"u}rgen Schmidhuber. 1997.
\newblock Long short-term memory.
\newblock \emph{Neural computation}, 9(8):1735--1780.

\bibitem[{Hudson and Manning(2019)}]{hudson2019gqa}
Drew~A. Hudson and Christopher~D. Manning. 2019.
\newblock \href {https://doi.org/10.1109/CVPR.2019.00686} {{GQA:} {A} new
  dataset for real-world visual reasoning and compositional question
  answering}.
\newblock In \emph{{IEEE} Conference on Computer Vision and Pattern
  Recognition, {CVPR} 2019, Long Beach, CA, USA, June 16-20, 2019}, pages
  6700--6709. Computer Vision Foundation / {IEEE}.

\bibitem[{Johnson et~al.(2017)Johnson, Hariharan, van~der Maaten, Fei{-}Fei,
  Zitnick, and Girshick}]{johnson2017clevr}
Justin Johnson, Bharath Hariharan, Laurens van~der Maaten, Li~Fei{-}Fei,
  C.~Lawrence Zitnick, and Ross~B. Girshick. 2017.
\newblock \href {https://doi.org/10.1109/CVPR.2017.215} {{CLEVR:} {A}
  diagnostic dataset for compositional language and elementary visual
  reasoning}.
\newblock In \emph{2017 {IEEE} Conference on Computer Vision and Pattern
  Recognition, {CVPR} 2017, Honolulu, HI, USA, July 21-26, 2017}, pages
  1988--1997. {IEEE} Computer Society.

\bibitem[{Kingma and Ba(2015)}]{kingma2017adam}
Diederik~P. Kingma and Jimmy Ba. 2015.
\newblock \href {http://arxiv.org/abs/1412.6980} {Adam: {A} method for
  stochastic optimization}.
\newblock In \emph{3rd International Conference on Learning Representations,
  {ICLR} 2015, San Diego, CA, USA, May 7-9, 2015, Conference Track
  Proceedings}.

\bibitem[{Lewis et~al.(2020)Lewis, Liu, Goyal, Ghazvininejad, Mohamed, Levy,
  Stoyanov, and Zettlemoyer}]{lewis2019bart}
Mike Lewis, Yinhan Liu, Naman Goyal, Marjan Ghazvininejad, Abdelrahman Mohamed,
  Omer Levy, Veselin Stoyanov, and Luke Zettlemoyer. 2020.
\newblock \href {https://doi.org/10.18653/v1/2020.acl-main.703} {{BART}:
  Denoising sequence-to-sequence pre-training for natural language generation,
  translation, and comprehension}.
\newblock In \emph{Proceedings of the 58th Annual Meeting of the Association
  for Computational Linguistics}, pages 7871--7880, Online. Association for
  Computational Linguistics.

\bibitem[{Pasupat and Liang(2015)}]{pasupat2015compositional}
Panupong Pasupat and Percy Liang. 2015.
\newblock \href {https://doi.org/10.3115/v1/P15-1142} {Compositional semantic
  parsing on semi-structured tables}.
\newblock In \emph{Proceedings of the 53rd Annual Meeting of the Association
  for Computational Linguistics and the 7th International Joint Conference on
  Natural Language Processing (Volume 1: Long Papers)}, pages 1470--1480,
  Beijing, China. Association for Computational Linguistics.

\bibitem[{Shaw et~al.(2018)Shaw, Uszkoreit, and
  Vaswani}]{shaw2018selfattention}
Peter Shaw, Jakob Uszkoreit, and Ashish Vaswani. 2018.
\newblock \href {https://doi.org/10.18653/v1/N18-2074} {Self-attention with
  relative position representations}.
\newblock In \emph{Proceedings of the 2018 Conference of the North {A}merican
  Chapter of the Association for Computational Linguistics: Human Language
  Technologies, Volume 2 (Short Papers)}, pages 464--468, New Orleans,
  Louisiana. Association for Computational Linguistics.

\bibitem[{Subramanian et~al.(2020)Subramanian, Bogin, Gupta, Wolfson, Singh,
  Berant, and Gardner}]{subramanian2020interpretability}
Sanjay Subramanian, Ben Bogin, Nitish Gupta, Tomer Wolfson, Sameer Singh,
  Jonathan Berant, and Matt Gardner. 2020.
\newblock Achieving interpretability in compositional neural networks.
\newblock In \emph{Association for Computational Linguistics (ACL)}.

\bibitem[{Suhr et~al.(2019)Suhr, Zhou, Zhang, Zhang, Bai, and
  Artzi}]{suhr2018corpus}
Alane Suhr, Stephanie Zhou, Ally Zhang, Iris Zhang, Huajun Bai, and Yoav Artzi.
  2019.
\newblock \href {https://doi.org/10.18653/v1/P19-1644} {A corpus for reasoning
  about natural language grounded in photographs}.
\newblock In \emph{Proceedings of the 57th Annual Meeting of the Association
  for Computational Linguistics}, pages 6418--6428, Florence, Italy.
  Association for Computational Linguistics.

\bibitem[{Talmor and Berant(2018)}]{talmor0218web}
Alon Talmor and Jonathan Berant. 2018.
\newblock \href {https://doi.org/10.18653/v1/N18-1059} {The web as a
  knowledge-base for answering complex questions}.
\newblock In \emph{Proceedings of the 2018 Conference of the North {A}merican
  Chapter of the Association for Computational Linguistics: Human Language
  Technologies, Volume 1 (Long Papers)}, pages 641--651, New Orleans,
  Louisiana. Association for Computational Linguistics.

\bibitem[{Talmor et~al.(2021)Talmor, Yoran, Catav, Lahav, Wang, Asai, Ilharco,
  Hajishirzi, and Berant}]{talmor2021multimodalqa}
Alon Talmor, Ori Yoran, Amnon Catav, Dan Lahav, Yizhong Wang, Akari Asai,
  Gabriel Ilharco, Hannaneh Hajishirzi, and Jonathan Berant. 2021.
\newblock Multimodalqa: Complex question answering over text, tables and
  images.
\newblock In \emph{International Conference on Learning Representations
  (ICLR)}.

\bibitem[{Vaswani et~al.(2017)Vaswani, Shazeer, Parmar, Uszkoreit, Jones,
  Gomez, Kaiser, and Polosukhin}]{vaswani2017attention}
Ashish Vaswani, Noam Shazeer, Niki Parmar, Jakob Uszkoreit, Llion Jones,
  Aidan~N. Gomez, Lukasz Kaiser, and Illia Polosukhin. 2017.
\newblock \href
  {https://proceedings.neurips.cc/paper/2017/hash/3f5ee243547dee91fbd053c1c4a845aa-Abstract.html}
  {Attention is all you need}.
\newblock In \emph{Advances in Neural Information Processing Systems 30: Annual
  Conference on Neural Information Processing Systems 2017, December 4-9, 2017,
  Long Beach, CA, {USA}}, pages 5998--6008.

\bibitem[{Wang et~al.(2020)Wang, Shin, Liu, Polozov, and
  Richardson}]{wang2020ratsql}
Bailin Wang, Richard Shin, Xiaodong Liu, Oleksandr Polozov, and Matthew
  Richardson. 2020.
\newblock \href {https://doi.org/10.18653/v1/2020.acl-main.677} {{RAT-SQL}:
  Relation-aware schema encoding and linking for text-to-{SQL} parsers}.
\newblock In \emph{Proceedings of the 58th Annual Meeting of the Association
  for Computational Linguistics}, pages 7567--7578, Online. Association for
  Computational Linguistics.

\bibitem[{Welbl et~al.(2018)Welbl, Stenetorp, and
  Riedel}]{welbl2018constructing}
Johannes Welbl, Pontus Stenetorp, and Sebastian Riedel. 2018.
\newblock \href {https://doi.org/10.1162/tacl_a_00021} {Constructing datasets
  for multi-hop reading comprehension across documents}.
\newblock \emph{Transactions of the Association for Computational Linguistics},
  6:287--302.

\bibitem[{Wolfson et~al.(2020)Wolfson, Geva, Gupta, Gardner, Goldberg, Deutch,
  and Berant}]{wolfson2020break}
Tomer Wolfson, Mor Geva, Ankit Gupta, Matt Gardner, Yoav Goldberg, Daniel
  Deutch, and Jonathan Berant. 2020.
\newblock \href {https://doi.org/10.1162/tacl_a_00309} {Break it down: A
  question understanding benchmark}.
\newblock \emph{Transactions of the Association for Computational Linguistics},
  8:183--198.

\bibitem[{Yang et~al.(2018)Yang, Qi, Zhang, Bengio, Cohen, Salakhutdinov, and
  Manning}]{yang2018hotpotqa}
Zhilin Yang, Peng Qi, Saizheng Zhang, Yoshua Bengio, William Cohen, Ruslan
  Salakhutdinov, and Christopher~D. Manning. 2018.
\newblock \href {https://doi.org/10.18653/v1/D18-1259} {{H}otpot{QA}: A dataset
  for diverse, explainable multi-hop question answering}.
\newblock In \emph{Proceedings of the 2018 Conference on Empirical Methods in
  Natural Language Processing}, pages 2369--2380, Brussels, Belgium.
  Association for Computational Linguistics.

\end{thebibliography}
\bibliographystyle{acl_natbib}

\appendix

\section{Appendices}
\label{sec:appendix}



\subsection{QDMR LF}
\label{apdx:LF}

Table \ref{tab:logical-form-props-and-args} shows the different operators, their properties and examples of LFs. Table \ref{tab:logical-form-props-indicators} shows terms that are used to identify the QDMR step operator's properties. We use the same lexicon from BREAK \cite{wolfson2020break} for detecting operators, extended with some specifications for numeric properties such as \prop{equals\_0}.

\begin{table*}
\small
\centering
\begin{tabular}{ m{2cm} m{2cm} m{2cm} m{9cm} }
\hline \textbf{Operator} & \textbf{PROP} & \textbf{ARG} & \textbf{Example} \\ \hline
\op{select}      & $\emptyset$         & \argn{sub}               & 
return cubes \\ &&& 
\op{SELECT}[](\argn{sub}=cubes) \\  
\hline
\op{filter}      & $\emptyset$         & \multirow{2}{2cm}{\argn{sub, condition}}         & 
return \#1 from Toronto \\ &&&
\op{FILTER}[](\argn{sub}=\#1, \argn{cond}=from Toronto) \\ 
\hline
\op{project}     & $\emptyset$         & \multirow{2}{2cm}{\argn{sub}, \argn{projection}}         & 
return the head coach of \#1 \\ &&&
\op{PROJECT}[](\argn{sub}=\#1, \argn{projection}=the head coach of) \\ 
\hline
\op{aggregate}   & \multirow{3}{2cm}{\prop{max, min, count, sum, avg}}      & \argn{arg}               &
return maximal number of \#1    \\&&&
\op{AGGREGATE}[\prop{max}](\argn{arg}=\#1)   \\ \\
\hline
\op{group}   & \multirow{3}{2cm}{\prop{max, min, count, sum, avg}}      & \argn{key, value}               &
return the number of \#2 for each \#1    \\&&&
\op{GROUP}[\prop{count}](\argn{key}=\#1, \argn{value}=\#2)   \\ \\
\hline
\op{superlative}   & \prop{max, min}      & \argn{sub, attribute}               &
return \#2 where \#3 is the lowest    \\&&&
\op{SUPERLATIVE}[\prop{min}](\argn{sub}=\#2, \argn{attribute}=\#3)   \\ 
\hline
\op{comparative}   & \multirow{6}{2cm}{\prop{equals, equals-[0/1/2], more, more-than-[0/1/2], less, less-than-[0/1/2]}}       & \multirow{3}{2cm}{\argn{sub, attribute, condition}}               &
return \#1 where \#2 is more than 100    \\&&&
\op{COMPARATIVE}[\prop{more}](\argn{sub}=\#1, \argn{attribute}=\#2, \argn{condition}=100)   \\ \\ \\ \\ \\
\hline
\op{comparison}   & \multirow{3}{2cm}{\prop{max, min, count, sum, avg, true, false}}      & \argn{arg}               &
return which is higher of \#1, \#2    \\&&&
\op{COMPARISON}[\prop{max}](\argn{arg}=\#1, \argn{arg}=\#2)   \\ \\ \\
\hline
\op{union}   & $\emptyset$     & \argn{sub}               &
return \#1, \#2    \\&&&
\op{UNION}[](\argn{sub}=\#1, \argn{sub}=\#2)   \\ 
\hline
\op{intersection}   & $\emptyset$     &  \multirow{2}{2cm}{\argn{intersect, projection}}               &
return parties in both \#2 and \#3    \\&&&
\op{INTERSECTION}[](\argn{intersect}=\#2, \argn{intersect}=\#3, \argn{projection}=parties)   \\ 
\hline
\op{discard}   & $\emptyset$     &  \argn{sub, exclude}               &
return \#1 besides \#2    \\&&&
\op{DISCARD}[](\argn{sub}=\#1, \argn{exclude}=\#2)   \\ 
\hline
\op{sort}   & $\emptyset$     &  \argn{sub, order}               &
return \#1 ordered by name    \\&&&
\op{SORT}[](\argn{sub}=\#2, \argn{order}=name)   \\ 
\hline
\op{boolean}   & \multirow{9}{2cm}{\prop{equals, equals-[0/1/2], more-than-[0/1/2], less-than-[0/1/2], and-true, and-false, or-true, or-false, if-exists}}      &  \multirow{2}{2cm}{\argn{sub, condition}}               &
return if \#1 is the same as \#2    \\&&&
\op{BOOLEAN}[\prop{equals}](\argn{sub}=\#1, \argn{condition}=\#2)   \\ \\ \\ \\ \\ \\ \\ \\ \\
\hline
\op{arithmetic}   & \multirow{2}{2cm}{\prop{sum, diff, multiply, div}}      &  \multirow{2}{2cm}{\argn{arg, left, right}}               &
return the difference of \#3 and \#4    \\&&&
\op{ARITHMETIC}[\prop{diff}](\argn{left}=\#3, \argn{right}=\#4)   \\
\hline
\end{tabular}
\caption{\label{tab:logical-form-props-and-args} LF operators, properties and arguments. Each QDMR step can be mapped to one of the above operators, where its LF consists of its operator, properties and arguments. The example column shows an example for such LF.}
\end{table*}

\begin{table}[t]
\small
\centering
\begin{tabular}{ m{2cm} m{1cm} m{3.5cm} }
\hline \textbf{Operator} & \textbf{PROP} & \textbf{Lexical entries} \\ \hline
\op{aggregate, comparison, group} & \prop{max} & max, most, more, last, bigger, biggest, larger, largest, higher, highest, longer, longest \\
\hline
\op{aggregate, comparison, group} & \prop{min} & min, least, less, first, fewer, smaller, smallest, lower, lowest,  shortest, shorter, earlier \\
\hline
\op{aggregate, comparison, group} & \prop{count} & count, number of, total number of \\
\hline
\op{aggregate, arithmetic, comparison, group} & \prop{sum} & sum, total \\
\hline
\op{aggregate, comparison, group} & \prop{avg} & avg, average, mean \\
\hline
\op{arithmetic} & \prop{diff} & difference, decline \\
\hline
\op{arithmetic} & \prop{multiply} & multiplication, multiply \\
\hline
\op{arithmetic} & \prop{div} & division, divide \\
\hline
\op{boolean, comparative} & \prop{equals} & equal, equals, same as \\
\hline
\op{boolean} & \prop{if-exists} & any, there \\
\hline
\op{comparative} & \prop{more} & more, at least, higher than, larger than, bigger than \\
\hline
\op{comparative} & \prop{less} & less, at most, smaller than, lower than \\
\hline
\op{superlative} & \prop{max} & most, biggest, largest, highest, longest \\
\hline
\op{superlative} & \prop{min} & least, fewest, smallest, lowest,  shortest, earliest \\
\hline

\end{tabular}
\caption{
\label{tab:logical-form-props-indicators}
Property lexicon. Tokens for detecting the properties of a QDMR step, for creating its logical form.
}
\end{table}
\subsection{LF-Based Evaluation (LF-EM)}
\label{sec:evaluation-LF-EM}

In \S\ref{sec:evaluation} we described a LF-based evaluation metric. Given a logical form of a QDMR, $Z$, the metric transforms it to a normalized form $Z_{norm}$ in the following way: (1) \emph{Normalize} the steps by removing unnecessary tokens and replace equivalents; (2) \emph{Merge} steps with their referrer; and (3) \emph{Reorder} the steps in a consistent order. Now we describe these steps more formally.

\paragraph{Normalize Steps} 
Let $z=\langle o, \rho, A \rangle$ be a logical form of step $s$, where $A=\{ \langle \eta^a, \tau^a \rangle \mid \eta^a \in \textit{PROP}_o, \tau^a \subseteq s \}_{a=1}^{|A|}$ is the named-arguments set. Note here $\tau$ is a set of tokens instead of sub-sequence of $s$. A \emph{normalization transformation} is a function $\mathcal{T}_{o,\rho}: s  \rightarrow \{ \emptyset \} \cup \sV_q \cup \sV_{store}$ mapping each token to an equivalent token out of the allowed vocabulary or to $\emptyset$ for removal. We denote by $\mathcal{T}_{o,\rho}(A) \coloneqq \{ \langle \eta, \{\mathcal{T}_{o,\rho}(\tau_1), \dots , \mathcal{T}_{o,\rho}(\tau_{|\tau|})\} \rangle \mid \langle \eta, \tau \rangle \in A\}$, 
i.e, applying a transformation on the named-arguments set is defined by applying it on each of the arguments tokens. The final normalized form of $l$ is given by applying multiple transformations on the arguments, $z^\text{norm}=\langle o, \rho,  \mathcal{T}_{o,\rho}^\textit{rep} \circ \mathcal{T}_{o,\rho}^\textit{aux} \circ \mathcal{T}_{o,\rho}^\textit{prop} (A)\rangle$.

$\mathcal{T}_{o,\rho}^{prop}$ removes the \emph{property lexicon entries} of properties $\rho$. This information is already given by $\rho$ and therefore is not needed to take a part in the arguments. Table \ref{tab:logical-form-props-indicators} shows the lexicon entries for each property.

$\mathcal{T}_{o,\rho}^{aux}$ removes uninformative tokens, such as prepositions. These $\sV_{aux}$ were added in the first place to allow continuous fluent sentences to be written.

$\mathcal{T}_{o,\rho}^{rep}$ maps a token to its \emph{representative token}. Recall \emph{BREAK} samples where annotated based on an allowed vocabulary, which is built on top of the question tokens variations and some additional ones. We define equivalence classes of \emph{break equivalent tokens} and set a representative for each class. Table \ref{tab:break-equivalance} shows some examples for these classes.

\begin{table}
\centering
\begin{tabular}{ m{2cm} m{4cm} }
\hline \textbf{Type} & \textbf{Equivalence Class} \\ \hline
Modifications   & cube, cubes, ... \\     
                & old, oldness, ... \\
                & taller, tall, ... \\
                & working, work, ... \\
\hline
Operational     & biggest, longest, highest, ... \\
\hline
Synonyms        & elevation, height  \\
                & 0, zero \\
                & ... \\
\hline
\end{tabular}
\caption{
\label{tab:break-equivalance} 
BREAK Equivalence Classes. (1) 
\emph{Modifications} - the same modifications of the question tokens that were used for creating BREAK annotation lexicon (e.g plural/singular form, nounify adjectives, lemmatize adjectives, lemmatize verbs); 
(2) \emph{Operational} equivalence induced from properties lexicon;
(3) Manually-defined \emph{Synonyms} lexicon; We mostly retrieve the final equivalence classes by merging classes that share some tokens.
}
\end{table}

\paragraph{Merge Steps} 
\label{sec:eval-logical-form:merge-rules}
The level of elaboration varies between annotators, leading to implicit steps, i.e, steps that are contained in other steps. This is especially common in \op{FILTER} and \op{PROJECT} steps. Therefore we offer a merging mechanism for grouping such decompositions. 

A merge rule is defined by $\langle o^{src}, \eta^{src}, o^{dst}, o^{out}, f_{\rho}, f_{\eta} \rangle $ where $o^{src}, \eta^{src}$ are the referrer step operator and argument name with the observed reference, $o^{dst}$ is the referred step operator, $o^{out}$ is the merged step operator and $f_{\rho}:\textit{PROP}_{o^{src}} \times \textit{PROP}_{o^{dst}} \rightarrow \textit{PROP}_{o^{out}}$, $f_{\eta}:\textit{ARG}_{o^{src}, \rho^{src}} \cup \textit{ARG}_{o^{dst}, \rho^{dst}} \rightarrow \textit{ARG}_{o^{out}, \rho^{out}}$ are mappings for the merged step properties and named arguments. The values for the arguments, $\tau$, are induced by merging the values of the arguments names that are mapped to the merged argument name.
\\
In particular, we use the following merging rules:
\begin{itemize}
  \item \texttt{project-sub $\rightarrow$ select = project} 
  Collapse select step that is referred by a project step as its subject.
  \begin{multline*}
      \langle o^{src}, \eta^{src}, o^{dst}, o^{out} \rangle =\\ \langle project, sub, select, project \rangle \\
      f_{\rho}(\rho^{src}, \rho^{dst}) = \rho^{src} \\
      f_{\eta}(a)=
      \begin{cases}
        a,  &\text{if } a \in ARG_{src} \\
        sub , &  \text{if } a \in ARG_{dst}
    \end{cases}  
  \end{multline*}
  
    \item \texttt{filter-sub $\rightarrow$ select = filter} 
  Collapse select step that is referred by a filter step as its subject.
  
    \item \texttt{filter-sub $\rightarrow$ filter = filter} 
  Collapse filter step that is referred by another filter step as its subject. This rule deals with filter chains, when a sequence of filters referred by each other with \emph{sub} argument, any order of them has the same meaning.
\end{itemize}

\paragraph{Reorder Execution Graph} 
In some cases there are multiple possible sequential orderings for the same execution graph, for example for parallel execution branches (Fig. \ref{fig:eval-normalization}). We reorder the graph by first splitting the steps into layers where each layer may refer to previous layers only, and then order within a layer lexicographically. Formally, let $\textit{REF}(z^i)\in[m]\setminus\{i\}$ be the references of step $z^i$. We define the \emph{degree} (layer) of $z^i$ by:
\begin{equation*}
 d(z^i) \coloneqq 
\begin{cases}
    0,  &\text{if } \textit{REF}(s)=\emptyset \\
    \max\limits_{j \in \textit{REF}(z^i)}{d(z^j)}+1, &  \text{otherwise}
\end{cases}   
\end{equation*}

Since QDMR execution graph is a DAG, $d(\cdot)$ is well defined. 
Let $\Delta_d = \{i \mid d(z^i)=d\}$ for $d \in [0..m]$. $rnk_{d(z^i)}(z^i)$ is the alphabet rank of $z^i$ in $\Delta_{d(z^i)}$, where $z^i$ textual representation is of the form $o^i[\rho^i_1, \dots, \rho^i_{|\rho^i|}](\eta^i_1=\tau^i_1, \dots, \eta^i_{|A^i|}=\tau^i_{|A^i|})$. The properties $\rho^i_1, \dots, \rho^i_{|\rho^i|}$ are sorted, and so the arguments $\eta^i_1=\tau^i_1, \dots, \eta^i_{|A^i|}=\tau^i_{|A^i|}$ first by the names $\eta$ and second by the values $\tau$. The textual representation of an argument value $\tau$ consists of alphabet ordered token, references first. Finally, the total rank of a step is given by $\langle d(z^i), rnk_{d(z^i)}(z^i) \rangle$, i.e primary order by degree and secondary order by in-layer rank.
\subsection{Experiments Parameters}
\label{apdx:experiments-parameters}

\paragraph{CopyNet-BERT}
The LSTM decoder has hidden size 768. We use a batch size of 32 and train for up to 25 epochs ($\sim$35k
steps) with beam search of size 5. 

\paragraph{Biaffine Graph Parser}
The POS embeddings are of size 100. The four FFNs consist of 3-layers with hidden size 300 and use ELU activation function. We use dropout of rate 0.6 on the contextualized encodings, and of rate 0.3 on the FF representations. We use a batch size of 32 and train for up to 80 epochs ($\sim$111k
steps).

\paragraph{Latent RAT} 
We stack 4 relation-aware self-attention layers on top of the contextualized encodings, each with 8 heads and dropout with rate 0.1. The FFNs for relation representation uses 3-layers with hidden size of 96, ReLU activation function and dropout rate of 0.1. We tie the layers, and multiply the graph loss by 100. The rest is identical to  the CopyNet-BERT configuration.

\paragraph{Optimization}
We used the Adam optimizer \cite{kingma2017adam} with the hyperparameters. The learning rate changes during training according to slanted triangular schema, in which it linearly increases from $0$ to $lr$ for the first $\textit{warmup\_steps}=0.06 \cdot \textit{max\_steps}$, and afterwards linearly decreases back to $0$.
We use learning rate of $1 \cdot 10^{-3}$,  and a separate learning rate of $5 \cdot 10^{-5}$ for the BERT-based encoder.

\onecolumn
\subsection{Error Analysis Examples}
\label{apdx:error-analysis}

Some examples for each error class from \S\ref{sec:analysis:error-analysis}. The gold decompositions are given on left, and the predictions are on the right.


\includegraphics[width=\textwidth]{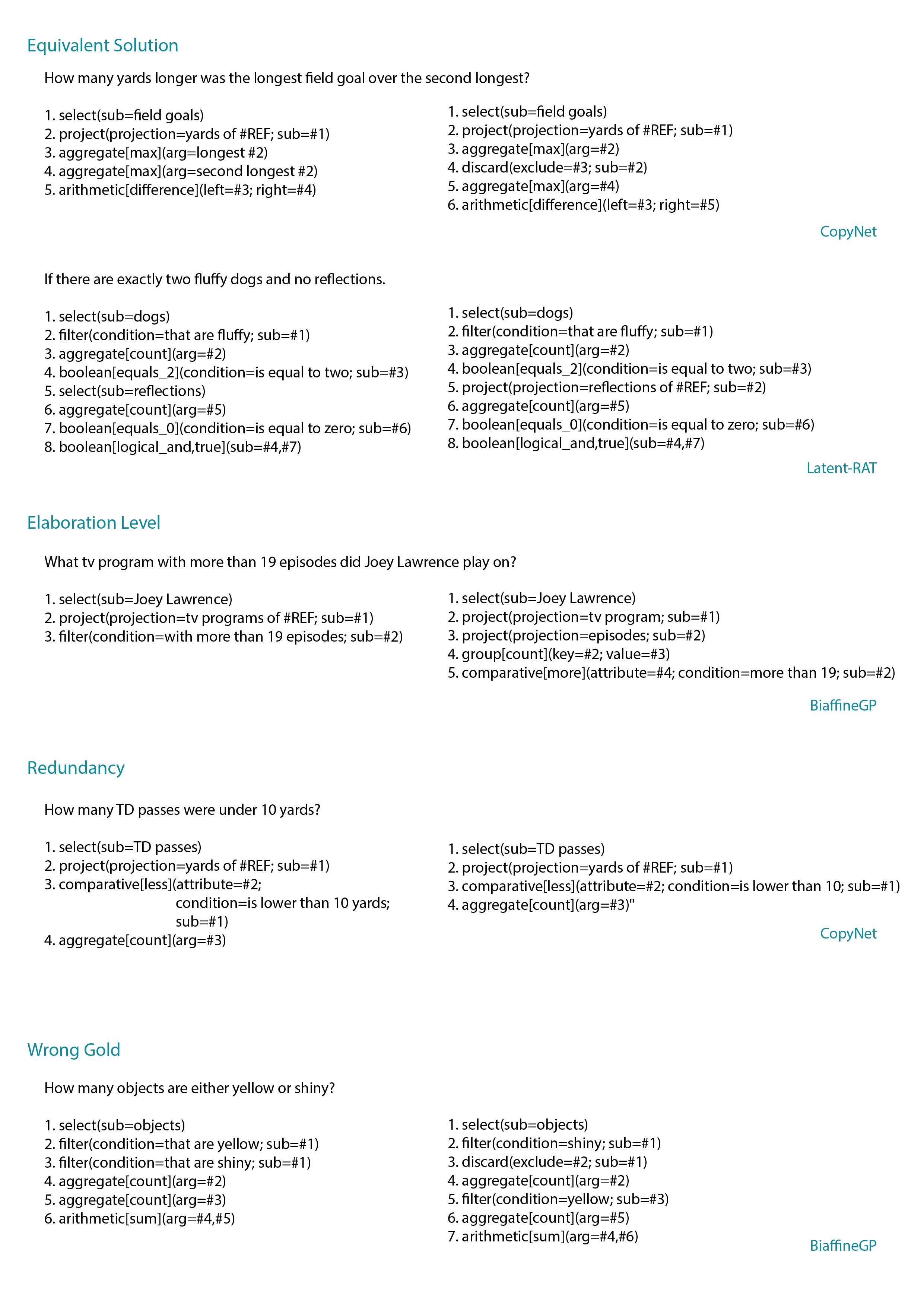}

\includegraphics[width=\textwidth]{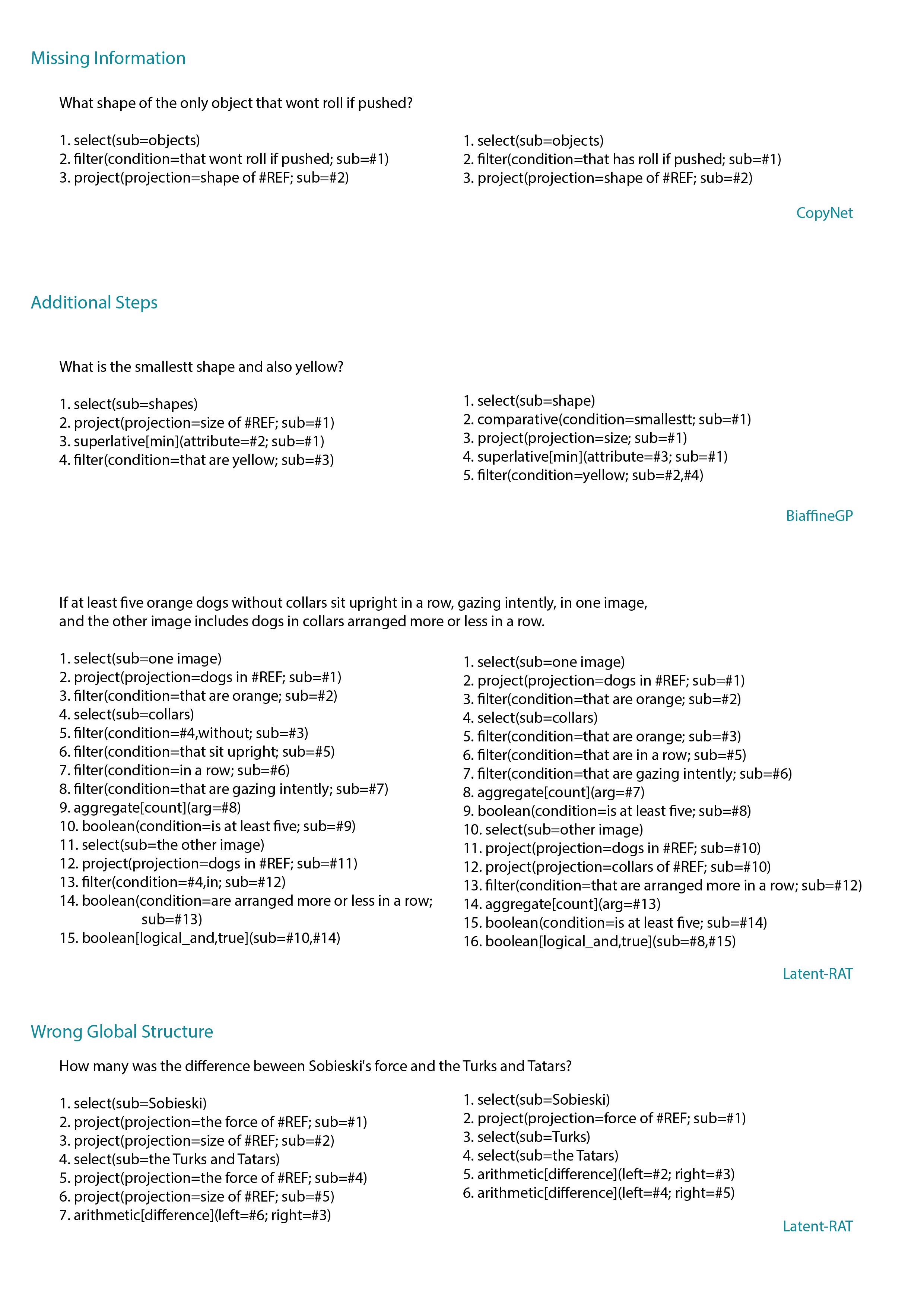}

\includegraphics[width=\textwidth]{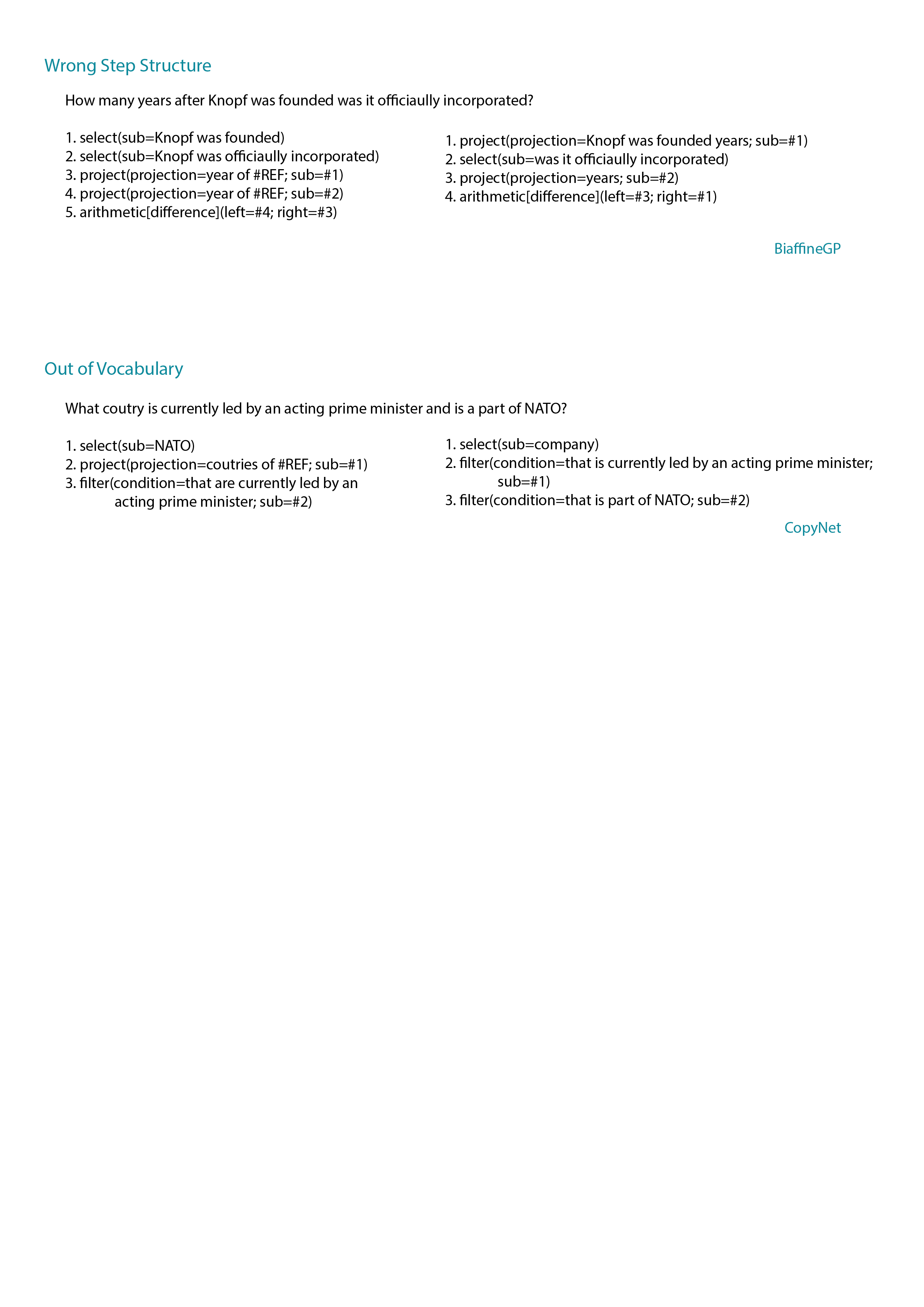}




\end{document}